\documentclass{article}

\usepackage{microtype}
\usepackage{graphicx}
\usepackage{subfigure}
\usepackage{booktabs} 
\usepackage{threeparttable}  

\usepackage{hyperref}
\hypersetup{colorlinks,allcolors=black}

\usepackage[accepted]{icml2025}

\usepackage{amsmath}
\usepackage{amssymb}
\usepackage{mathtools}
\usepackage{amsthm}

\usepackage[capitalize,noabbrev]{cleveref}

\theoremstyle{plain}

\theoremstyle{definition}

\theoremstyle{remark}

\usepackage[textsize=tiny]{todonotes}

\usepackage{xcolor}
\usepackage{enumitem}
\usepackage{xspace}
\usepackage[normalem]{ulem}

\usepackage{algorithm}

\usepackage{algpseudocode}
\usepackage{xcolor} 
\algrenewcommand\algorithmiccomment[1]{\hfill\textcolor{gray}{// #1}}

\usepackage[most]{tcolorbox}
\usepackage{adjustbox}  
\usepackage[normalem]{ulem}

\newif\ifshownotes
\shownotestrue  

\ifshownotes
  \newenvironment{notelist}{
    \color{gray}
    \begin{itemize}[
      label=--,leftmargin=2em, itemsep=0pt, topsep=0pt, parsep=0pt, partopsep=0pt
    ]
  }{
    \end{itemize}
  }
\else
  
  \usepackage{verbatim} 
\fi

\newcommand{\ouralgo}{\textsc{UIExplore-AlGo}\xspace}
\newcommand{\ourbench}{\textsc{UIExplore-Bench}\xspace}

\newcommand{\uniqueElements}{\textsc{UI‑Functionalities Observed}\xspace}
\newcommand{\abruniqueElements}{\textsc{UFO}\xspace}
\newcommand{\abrnormUniqueElements}{\textsc{hUFO}\xspace}

\newcommand{\uniqueFunctionalitiesTested}{\textsc{UI‑Functionalities Tested}\xspace}
\newcommand{\abruniqueFunctionalitiesTested}{\textsc{UFT}\xspace}

\icmltitlerunning{UI Exploration ICML 2025}

\begin{document}

\twocolumn[
\icmltitle{Toward Autonomous UI Exploration: The UIExplorer Benchmark}



\icmlsetsymbol{equal}{*}

\begin{icmlauthorlist}
\icmlauthor{Andrei Cristian Nica}{equal,UiPath}
\icmlauthor{Akshaya Vishnu Kudlu Shanbhogue}{equal,UiPath}
\icmlauthor{Harshil Shah}{UiPath}
\icmlauthor{Aleix Cambray}{UiPath}
\icmlauthor{Tudor Berariu}{UiPath}
\icmlauthor{Lucas Maystre}{UiPath}
\icmlauthor{David Barber}{UiPath,UCL}
\end{icmlauthorlist}

\icmlaffiliation{UiPath}{UiPath}
\icmlaffiliation{UCL}{UCL, London, UK}

\icmlcorrespondingauthor{Andrei Cristian Nica}{andrei.nica@uipath.com}

\icmlkeywords{Machine Learning, ICML}

\vskip 0.3in
]



\makeatletter
\def\Hy@Warning#1{}
\makeatother

\printAffiliationsAndNotice{\icmlEqualContribution} 

\begin{abstract}

Autonomous agents must know how to explore user interfaces (UIs) for reliable task solving, yet systematic evaluation of this crucial phase is lacking. We introduce \ourbench, the first benchmark explicitly dedicated to UI exploration. The benchmark evaluates agents with either Structured mode (granting access to layout information like DOM trees) or Screen mode (relying on GUI-only observations such as screenshots and human-like mouse/keyboard interactions) across three levels in a standardized GitLab sandbox environment. We formalize exploration as the process of maximizing the set of actionable UI components discovered and propose a metric, human-normalized \uniqueElements (\abrnormUniqueElements), to quantify the effectiveness of exploration. 
Our results show that \ouralgo achieves the leading mean \abrnormUniqueElements scores, reaching up to 77.2\% of human performance in Structured mode and 59.0\% in Screen mode at 2,000 steps, particularly excelling at the Sparse level. The results highlight the relevance of our benchmark, as current agents show a substantial performance gap compared to 1 hour of human expert exploration, indicating ample room for future advancements. We publicly release the benchmark environment, an exploration dataset, and an evaluation suite to catalyze research into efficient UI exploration strategies and their downstream applications, such as experience-driven task completion and automated training data generation.

\end{abstract}

\section{Introduction}
\label{sec:intro}


The pursuit of truly autonomous agents capable of interacting with the digital world hinges on their ability to understand and navigate user interfaces. Imagine a robotic process automation bot landing on a freshly upgraded payroll dashboard, with new features and user interface (UI) improvements. In such a scenario, and countless others, the agent cannot rely solely on pre-programmed instructions or static UI knowledge. Frequent UI changes and the sheer diversity of applications require modern agents to actively explore and discover available functionalities. This exploration is not merely a preliminary step; it is a fundamental learning process \cite{schmidhuber2008driven, schmidhuber1991possibility, thrun1992efficient}.

We posit that efficient exploration is paramount to achieving reliable task execution and genuine autonomy in dynamic digital environments. At a high level, the performance of an UI agent is determined by two factors. First, its awareness of the UI space, such as its understanding of what functionalities exist and where interactions lead. Second, the agent's ability to solve tasks successfully, \emph{conditioned on its knowledge of the UI space}. This distinction is important: for example, an agent might have access to the knowledge required to complete a task but might fail to precisely locate the right UI element. Measuring what situations an agent has been exposed to or not, independently of its ability to solve tasks, is valuable, as it helps to disentangle these two performance factors. Robust UI exploration can provide a mechanism to explicitly address and improve both factors. It can address the first by maximizing knowledge of the UI space instead of success in specific tasks. It can address the second by encouraging agents to minimize the discrepancy between the predicted consquences of an action and the actual observed outcome.

At a more granular level, the design of effective exploration policies provides the following benefits.
(i)~Enhancing unattended data generation by efficiently discovering the functional space of the UI. This practice is already used to gather training experiences \cite{su2025learn, fan2025guibee, qi2024webrl, bai2024digirl}, and more efficient methods are highly valuable.
(ii)~Uncovering the ``unknowns'' of the agent's underlying predictive model and its misalignments \cite{fan2025guibee}. This could be especially beneficial for visual computer-use agents, which often fall short in grounding UI functionality \cite{agashe2025agent} and often misinterpret the purpose or effects of the interface elements \cite{zhang2025characterizing}. A dedicated benchmark can encourage research into rewarding agents for such information-seeking behavior.
(iii)~Enabling, through structured exploration, the creation of persistent databases of application-specific memory. This could offer a scalable solution to improve generalist agents with deep contextual understanding and opens an interesting research direction.
Driven by these three benefits, our work addresses the need for principled methods to measure efficient UI coverage and guide the design of effective policies.

Despite the recognized importance of exploration, there is a significant gap in systematic evaluation. Current benchmarks for UI agents predominantly entangle exploration with task-specific rewards \cite{zhou2023webarena, chezelles2024browsergym, drouin2024workarena, boisvert2025workarenacompositionalplanningreasoningbased}. Consequently, empirical comparisons make it difficult to disentangle the contributions of an agent's exploratory prowess from its ability to exploit learned knowledge for a specific goal. Most existing research either conflates these two aspects or focuses on exploration merely as a mechanism to gather experiences to improve exploitation, often acknowledging that more sophisticated exploration strategies would be beneficial \cite{su2025learn}.

To bridge this gap, we introduce \ourbench, the first benchmark designed for systematic assessment of UI functionality exploration. \ourbench offers a standardized GitLab sandbox environment (adapted from WebArena \cite{zhou2023webarena}) with three difficulty levels, for agents operating in Structured or Screen mode. We formalize exploration as the process of maximizing the discovery of unique functionalities offered by a system, captured by our proposed metric: \uniqueElements. Designed to surface key exploration challenges, \ourbench — together with its publicly released exploration dataset - aims to promote and standardize research on agents that explore more efficiently. In addition, the fixed dataset enables fairer comparison of exploitation strategies and clearer attribution of performance gains to either improved exploration or better use of prior knowledge.

\noindent Our contributions are fourfold:
\begin{itemize}[leftmargin=*, itemsep=1pt, topsep=1pt, partopsep=0pt, parsep=0pt]
    \item We publicly release \textbf{\ourbench}, a benchmark environment with three difficulty levels based on a GitLab sandbox, supporting both Structured and Screen modes.
    \item We introduce two novel UI-centric metrics that capture functionality discovery (\uniqueElements) and exploration efficiency (\uniqueFunctionalitiesTested), as well as a lightweight evaluation suite.
    \item We develop  \ouralgo, a Go-Explore-inspired \cite{ecoffet2019go} novelty-driven agent designed for structured UI exploration and memory construction, and compare its performance to baseline agents: random, BFS/DFS and GUI-Bee$\ddagger$ .
    \item We publicly release an open exploration dataset gathered with \ourbench, enabling downstream research on experience-driven task completion, metric learning, and more.
\end{itemize}


\section{Related Work}
\label{sec:related}

\textbf{Exploration in Reinforcement Learning.} Effective exploration in reinforcement learning (RL), especially with sparse rewards, remains a significant research challenge. Common paradigms include count-based methods \cite{bellemare2016unifying}, curiosity-driven intrinsic rewards utilizing prediction errors \cite{pathak2017curiosity, burda2018large}, and information-theoretic strategies such as VIME \cite{houthooft2016vime}. Episodic memory-based approaches \cite{badia2020never} and hierarchical structures \cite{steccanella2020hierarchical, nica2022paradox} have further enabled structured exploration in complex environments. Our proposed algorithm, \ouralgo, builds upon this literature by integrating a structured and hierarchical search strategy inspired by Go-Explore \cite{ecoffet2019go, ecoffet2021first, lu2024intelligent} that uses novelty and preference-based prioritization to efficiently explore user interfaces.

\textbf{Computer-Use Agent Benchmarks}. A growing set of benchmarks has been proposed to evaluate agents operating digital interfaces, including web-based environments like WebArena \cite{zhou2023webarena}, BrowserGym \cite{chezelles2024browsergym}, MiniWoB \cite{10.5555/3305890.3306005} and OpenWebVoyager \cite{he2024openwebvoyager}, as well as productivity-focused setups such as WorkArena \cite{drouin2024workarena} and WorkArena++ \cite{boisvert2025workarenacompositionalplanningreasoningbased}. These benchmarks primarily target the completion of goal-conditioned tasks and often conflate exploration with exploitation, making it difficult to isolate and study exploration behaviors in depth. Vision-centric environments such as OSWorld \cite{xie2024osworld} further emphasize the challenges of agents interacting through human-like modalities, e.g., mouse, keyboard, and screen input, without access to privileged structural information. However, they still lack dedicated tools and metrics to systematically evaluate exploration as a standalone capability. In contrast, \ourbench introduces a standardized benchmark explicitly designed to assess UI functionality exploration, providing both Structured (DOM-based) and Screen (GUI-based) modes, and metrics focused on functional discovery and exploration efficiency. This allows for a clearer analysis of exploratory competence - an essential precursor to robust and general-purpose computer-use agents.

\textbf{Computer-Use Agents.} Recent state-of-the-art computer-use agents leverage diverse modalities and architectures. General purpose frameworks such as OpenAI's Computer-Using Agent (CUA \cite{openai2025cua}, Simular Agent S2 \cite{agashe2025agent}, and Claude 3.5 Sonnet Compute-use \cite{anthropic2024computeruse} demonstrate strong visual grounding and reasoning capabilities across web, desktop, and mobile tasks. Enterprise-focused agent frameworks like Microsoft AutoGen \cite{wu2023autogen} and IBM CUGA \cite{marreed2025towards} leverage machine learning agents to automate the resolution of tasks and workflows. 

\textbf{LLM-Powered Agents and UI Exploration.} Several recent works explicitly or implicitly leverage exploration within their approaches. MobileGPT \cite{lee2024mobilegpt} and AppAgent \cite{li2024appagent} explore app interfaces to build structured knowledge bases and reusable task modules. Symbiotic Cooperation for Web Agents \cite{zhang2025symbiotic} integrates exploration with small-model distillation loops, and Explorer \cite{pahuja2025explorer} employs large-scale LLM-driven trajectory synthesis. WebWalker \cite{wu2025webwalker} introduces a multi-agent system where an explorer navigates multi-hop web interfaces while a critic evaluates information sufficiency, aiming to improve deep web exploration. LATS \cite{zhou2023language}, Tree Search for Language Model Agents \cite{koh2024tree}, and ExACT \cite{yu2024exact} utilize advanced tree-search methods to enhance targeted traversal and exploration efficiency. Learn-by-interact \cite{su2025learn} further generates exploration data through self-instructed tasks. Another example is GUI-Bee \cite{fan2025guibee}, which explicitly uses exploration to uncover actionable UI elements and learn from interaction data; it combines a Q-learning framework with a structured graph over the action space to guide exploration in a sample-efficient manner. Unlike these methods that mainly use exploration instrumentally, \ourbench prioritizes systematic measurement and improvement of exploration itself, establishing its distinct value for progress in computer use agent research.

By explicitly isolating exploration and introducing rigorous metrics, \ourbench not only advances exploration evaluation but provides foundational insights useful for improving downstream tasks such as retrieval-augmented generation (RAG)-conditioned planning and model introspection \cite{yu2024exact, koh2024tree, zhou2023language}.


\section{\ourbench Benchmark Suite}
\label{sec:benchmark}

\ourbench is designed to systematically evaluate the UI exploration capabilities of autonomous agents within a standardized and complex sandbox environment derived from the \texttt{WebArena} GitLab setup. GitLab was chosen due to its rich interactivity, multi-layered navigation structure, and clear metric computation facilitated by privileged DOM data. It also offers extensive opportunities for agents to create and manipulate content (such as projects, issues, and files), making it ideal for evaluating goal-directed exploration and interactive discovery. This environment serves as a controlled yet challenging testbed to measure agents' abilities to explore and discover interactive functionalities. Although initially designed as a web-based benchmark, we propose extending it to the vision-only, general-purpose computer interaction setting, facilitating evaluation beyond web applications and towards broader GUI-based interaction scenarios.

\subsection{Environment Design}
\label{Environment-Design}

We structure the benchmark into \textbf{three distinct levels}, corresponding to three different initial states.
Each level introduces unique challenges and constraints to the exploration problem.
 
\textbf{Abundant Level}. The agent starts in a logged-in state with extensive repositories and visible interactive elements (as per the original WebArena setup). This environment tests agents' abilities to effectively navigate and distinguish actionable components amidst abundant but potentially distracting UI elements.

\textbf{Moderate Level}. The agent remains logged in, but faces a reduced, repository-free UI skeleton, minimizing duplicative interactions. Exploration at this level emphasizes efficient coverage of essential UI functionality without the complexity of content-heavy pages.

\textbf{Sparse Level}. Simulates a fresh GitLab instance, without users and repositories, requiring the agent to initiate exploration from a logged-out state. Although minimal, there are still a few functionalities to explore even without registering a user. This scenario assesses the agent’s capacity to bootstrap exploration from minimal initial state information. It also assesses the agent's ability for open-world exploration.

To better align with established efforts in the computer-use community, we formalize \textbf{two distinct interaction modes}.

\textbf{Structured Mode}: The agent interacts with environments through the Document Object Model (DOM), perceiving the hierarchical structure of elements, and performing high-level actions (clicking, form filling, hovering) identified via DOM attributes. This mode leverages the semantic web structure, exemplified by benchmarks such as WebArena \cite{zhou2023webarena}.

\textbf{Screen Mode}: The agent relies solely on visual inputs, perceiving the UI as rendered images without having access to the underlying DOM structures. Actions are low-level, including mouse movements, pixel-coordinate clicks, and keyboard presses, emphasizing visual understanding and mimicking human interactions. This mode aligns with benchmarks such as OSWorld \cite{xie2024osworld}.

Additional details on observation and action spaces for these interaction modes are provided in section \ref{sec:spaces}.
Each level includes guardrails to ensure a self-contained exploration experience, preventing agents from accessing external sites or performing disruptive actions like accidental logouts. External links or pop-ups are sandboxed via a boundary HTML page (detailed in Appendix~\ref{app:env-details}).

\subsection{Exploration Protocol}
\label{sec:protocol}

Exploration episodes begin with an exploration agent starting at the GitLab landing page, with session storage cleared to ensure a consistent starting state. The agent is evaluated over a single trajectory that spans 2,000 environment actions (atomic actions),  reflecting a realistic and extended interaction scenario where environment changes persist throughout the exploration. This single-trajectory protocol ensures that agents are evaluated on sustained exploratory behaviors rather than repeated short-term interactions.

Structured and Screen modes support navigation actions such as \texttt{goto}, \texttt{go\_back}, and \texttt{go\_forward}. Explicit support for these actions aligns the benchmark with real-world applications that typically provide deep linking, history navigation, or saved states, thus enhancing practicality. Future extensions of Screen mode may include environments restricting such navigational aids, offering alternative exploration challenges. Details of the exploration protocol, including reset semantics, are provided in Appendix~\ref{app:env-details}.

\subsection{Metrics}                            
\label{sec:metrics}

An exploration agent's primary goal is to identify the functionalities exposed by an application in order to exploit it later. Our primary metric is the number of UI-functionalities observed at time $T$, denoted as $\text{UFO@}T \doteq \lvert \cup_{t=1}^T F_t \vert$, where $F_t$ is the set of distinct functionalities revealed at step $t$. These elements are grouped not only by semantic intent but also by their implementation in the UI structure; for instance, visually distinct ``Home'' buttons may count as separate functionalities, while all repository links map to a shared class such as $view\_repo$. Examples of functionalities can be found in Fig.~\ref{fig:homepage-metric-groupings}. The complete formalism and definition of functionality is in Appendix~\ref{app:metric-formalism}. In Structured mode, all functionalities accessible in the DOM are considered in the computation of \abruniqueElements. However, in Screen mode, only the functionalities accessible from the screen (viewport) are considered.

We also normalize \abruniqueElements with respect to the average number of \abruniqueElements discovered by human exploration agents in 1 hour. We denote this human-normalized version as \abrnormUniqueElements. In order to normalize the score, we collected exploration trajectories from human participants (Appendix~\ref{app:human-evaluation}) in the Abundant level. The normalization constants for the Structured mode and the Screen mode can be found in Table~\ref{tab:human_scores}. Our secondary metric, UI‑functionalities tested (UFT@T), counts the distinct functionalities that an exploration agent acted upon, as a percent of the total number of steps. \abruniqueFunctionalitiesTested acts as a measure of exploration efficiency and is in the range $[0, 1]$. 

We chose not to adopt per-observation count metrics, such as the Depth-fixed DOM Diversity Counts (D3C) introduced in GUI-Bee \cite{fan2025guibee}, as they treat all observations equally. Our decision was based on the following considerations.
\((i)\) Different observations could differ significantly in the number of functionalities they contain. Treating all observations as equally informative fails to capture this variability and undervalues richer states that offer access to multiple functionalities.
\((ii)\) D3C includes structured elements regardless of its visibility on the screen. Due to this, D3C is only applicable for Structured mode and is not applicable in Screen mode.
\((iii)\) Minor changes in the DOM structure can significantly affect the score (e.g. user's ``theme'', see Appendix~\ref{app:d3c-analysis}).


\begin{figure}[t]
    \centering
    \fbox{
        \includegraphics[width=0.9\linewidth, clip=true, height=4.5cm]{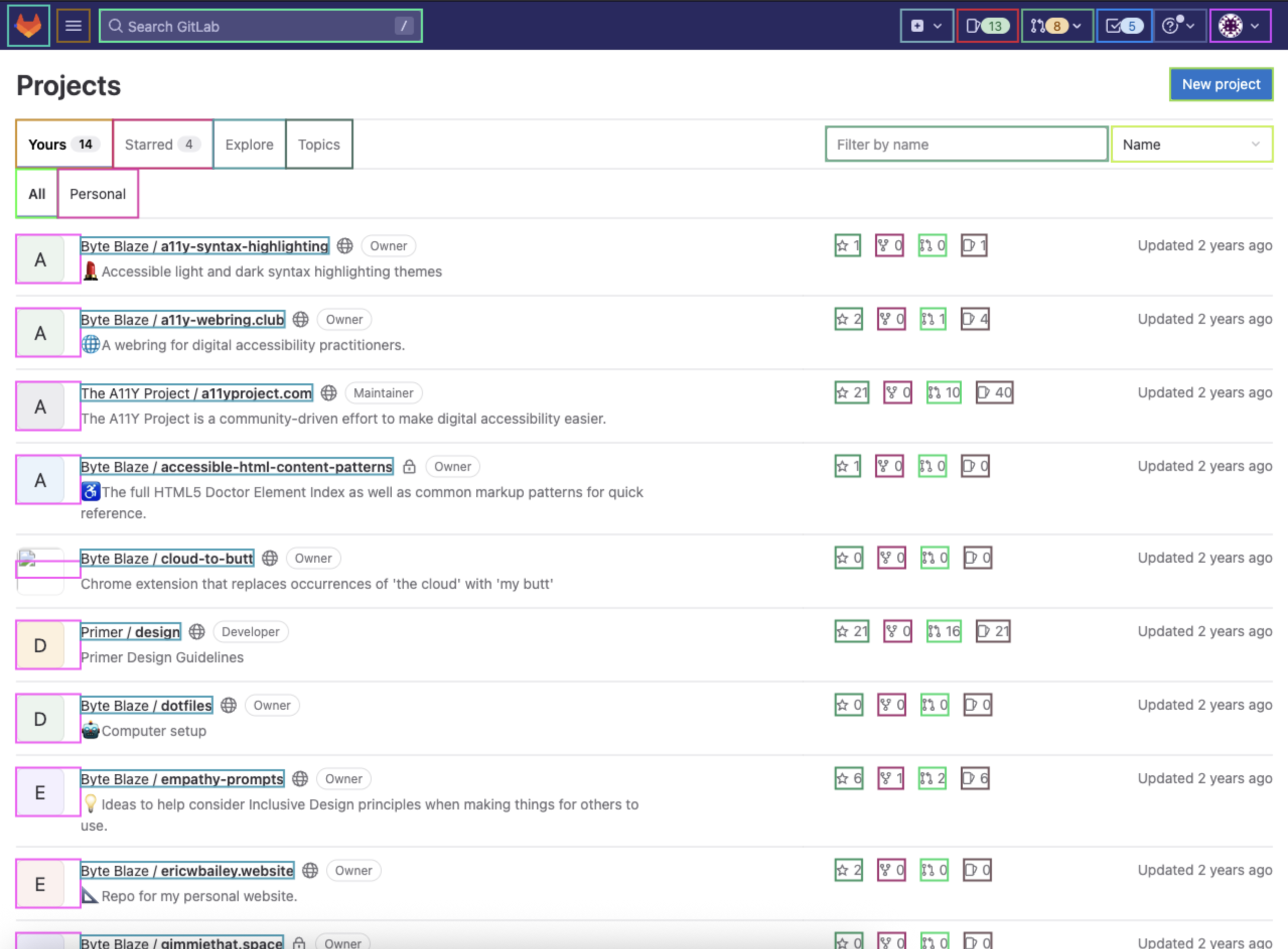}
    }
    \caption{Functionality groupings in homepage. Boxes of the same color are grouped under the same functionality}
    \label{fig:homepage-metric-groupings}
    \vspace{-12pt} 

\end{figure}
 
\subsection{Observation \& Action Spaces}
\label{sec:spaces}

We distinguish between two agent modalities with distinct observation and action spaces: Structured and Screen mode.

\textbf{Structured Mode.} Agents have access to a structured representation of the interface, such as the DOM or AXTree. Each element is identified via a unique identifier (e.g., \texttt{bid}), along with semantic metadata like tag, role, label, and visibility. This enables direct access to high-level functionality. The action space consists of 11 semantic actions including \texttt{click(bid)}, \texttt{fill(bid, value)}, \texttt{goto(url)}, and \texttt{keyboard\_press(key)}. These are discrete, parameterized actions grounded in the DOM structure.

\textbf{Screen Mode.} Agents receive only raw pixel observations from the UI, rendered as 1280×720 RGB images. Structural information is not available. The action space includes 18 low-level operations, such as \texttt{mouse\_move(x, y)}, \texttt{mouse\_click(x, y)}, \texttt{keyboard\_type(text)}, and \texttt{scroll(dx, dy)}. These actions operate in pixel coordinates and approximate human-like GUI interactions.

In both modes, the environment also provides a natural language instruction that defines the available actions and their parameters, facilitating integration with language models.

\section{Agents}
\label{sec:algorithms}

To provide comprehensive reference points for various methodological approaches, we evaluated a range of agents. These include random baselines, heuristic search methods, established techniques from existing literature, and approaches that take advantage of oracle information. The latter, particularly for Screen mode, helps establish an understanding of potential performance ceilings when, for instance, perfect bounding-box information for interactive elements is assumed. This variety allows for a clearer delineation of challenges and capabilities specific to UI exploration.

A key consideration in our evaluation is the distinction between agents operating in the Structured mode versus Screen mode, each with its own observation and action space constraints (see Section \ref{sec:spaces}).

\subsection{Baselines}
\label{Baselines}

We include several baseline agents to contextualize the performance of our proposed method and highlight specific exploration challenges. In the case of Screen mode, many agents such as GUI-Bee rely on the UI context grounding phase as part of agent execution. This phase is responsible for the extraction of potential actions from screenshots. In order to isolate it's effects on exploration and help ensure a fair comparison of their search strategies rather than their visual grounding capabilities, we provide certain agents access to perfect information (such as bounding boxes) of interactive elements on the webpage. Such agents are said to possess ``oracle'' information.

\textbf{Random Agent}: This agent serves as a fundamental baseline. In Structured mode, the agent performs one of the following actions uniformly at random: \texttt{scroll\_up}, \texttt{scroll\_down}, \texttt{go\_back}, and \texttt{click} (for each of the DOM elements). In Screen mode, the agent chooses one of the following actions: \texttt{scroll\_up}, \texttt{scroll\_down}, \texttt{go\_back}, and \texttt{click} on a random pixel with weights $0.09$, $0.09$, $0.02$, and $0.80$, respectively.

\textbf{Heuristic Random Agent}: This agent assumes access to the perfect bounding-box information of interactive DOM elements on the webpage. The agent selects an action uniformly at random from the list of \texttt{click} actions for each of the interactive DOM elements, along with \texttt{scroll\_up},  \texttt{scroll\_up} and  \texttt{go\_back}. In the Screen mode, the agent limits the click actions to only the elements visible on the screen.

\textbf{URL-Space Search (BFS \& DFS)}: These baselines operate in the Structured mode by extracting hyperlinks from the DOM tree. We apply both Breadth-First Search (BFS) and Depth-First Search (DFS) strategies to systematically explore the discovered URL space.

\textbf{GUI-Bee$\ddagger$}: As a baseline from the literature, our implementation adapts the GUI-Bee algorithm \cite{fan2025guibee}. It extracts click, scroll, and go-back actions based on Q-value estimates derived from a Q-table. To maintain simplicity and focus on the core exploration dynamics, our version omits the model-based Q-value reestimation step. This agent is primarily designed for and evaluated in the Screen mode.

\textbf{Human Evaluation}: To provide a critical reference point, we include performance data from human evaluators. Three participants with varying levels of familiarity with the GitLab instance were involved: a proficient user very familiar with the specific GitLab instance, a developer familiar with the underlying technology, and an amateur non-technical user. Participants were asked to identify all the functions that Gitlab provides, and their interactions were captured for 1 hour. More details can be found in Appendix~\ref{app:human-evaluation}.

\begingroup
\begin{algorithm}[H]
\caption{\ouralgo: Exploration Loop}
\label{alg:uiexplorer}
\begin{algorithmic}[1]
\Require start state $s_0$, $N_M$ number of macro-actions, $N_A$ number of atomic-actions
\State $G \gets \{\}$  \Comment{knowledge graph with visited states}
\State $s \gets s_0$
\While{\textbf{True}}
    \For{$i = 1$ \textbf{to} $N_M$}
        \State $(s_d, A) \gets$ describe\_state\_and\_macro\_actions($s$)
        \State $a_M \gets$ choose\_novel\_macro\_action($s$, $A$, G)
        \While{\textbf{not} finished($a_M$) \textbf{and} step $< N_A$}
            \State $a_A \gets$ act\_agent.execute($a_M$)
        \EndWhile
        \State $G \gets G \cup \{s,s_{d},A, a_{A_i..}\}$

    \EndFor
    \State $s \gets$ goto(choose\_frontier\_macro\_action($G$))
\EndWhile
\end{algorithmic}
\end{algorithm}
\vspace{-16pt} 
\endgroup

\subsection{UIExplore-AlGo}
\label{UIExplorer}

Our algorithm is a Go-Explore–style framework that treats UI exploration as goal-directed map-building. At every step we prompt GPT-4o to produce a compact natural-language description of the state and possible next actions. These textual outlines, rather than raw screenshots, become the keys in our novelty table; this makes the scorer resilient to visual drift (e.g., GitLab's theme switch) while still rewarding unseen functionality.

Exploration proceeds hierarchically.  
GPT-4o first suggests macro-actions (e.g., “create project”) that can unlock whole new regions of the state space but would be prohibitively slow to discover with atomic actions alone. Within a macro-action, a smaller model (Claude) executes up to $N_A$ atomic steps, giving us the best of both worlds: deliberate high-level search as well as fast, low-level control.

A candidate macro-action is ranked by a tri-objective score,
\((i)\) predicted novelty of its endpoint,  
\((ii)\) dissimilarity from past trajectories (to avoid rabbit holes), and 
\((iii)\) an importance prior from GPT-4o that favours actions likely to unlock potentially new areas of functionality.  
This mixture biases the agent toward cumulative, long-horizon payoff instead of one-step curiosity.

Although we can issue a \texttt{goto} to return to a previously visited state, the restored environment may not perfectly reproduce the original (e.g., form inputs may be lost). To mitigate this, the agent executes several macro-actions before “teleporting” again, amortizing the cost of resets and enabling batch exploration of multiple potential frontiers.

Algorithm~\ref{alg:uiexplorer} summarises the loop; full prompts and an annotated trajectory appear in Appendix~\ref{app:uiexplorer}.


\section{Experimental Setup}
\label{sec:experiments}

 Our experimental setup evaluates six agents across three GitLab environment levels (Abundant, Moderate, and Sparse) and two interaction modalities: Structured (with DOM access) and vision-based Screen mode. We use \abrnormUniqueElements as the primary metric, normalized against human reference scores specific to each mode and environment level. Each agent performs a single exploration trajectory of up to 2,000 atomic actions. We also report a multi-seed analysis for the Screen-Moderate setup detailed in Section~\ref{sec:ablation}.
 

We leveraged the BrowserGym framework with Playwright for environment control and Docker for containerization. Performance is normalized by atomic action count to ensure agent-centric fairness, isolating policy efficiency from hardware or network latencies.

For our proposed agent, \ouralgo, the state and macro-action descriptions are generated by \textit{gpt-4o-2024-11-20}, while the action execution relies on \textit{claude-3-5-sonnet-v2@20241022}, chosen for its Screen coordinate prediction capabilities. \ouralgo uses a maximum of 6 and 12 atomic actions per macro-action in the Structured and Screen modes, respectively, reflecting the differing action complexities. 


\section{Results}
\label{sec:results}

This section presents key findings from our benchmark, including agent performance, ablation studies, and qualitative insights into exploration patterns and challenges.

\subsection{Exploration Performance}

\begin{figure*}[t]
  \centering
  \includegraphics[height=6.5cm]{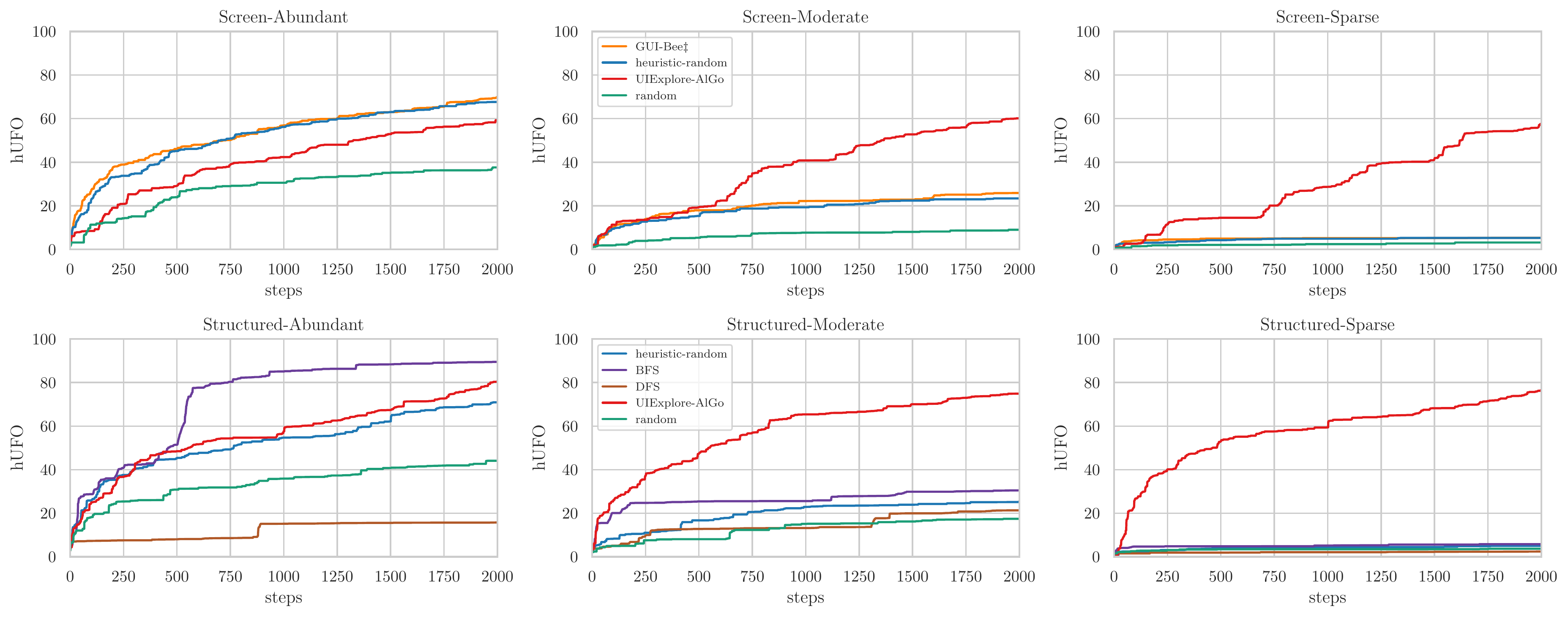}
  \vspace{-12pt} 
  \caption{Progress of human-normalized UI Functional Discovery (\uniqueElements) over exploration steps (atomic-actions). Top row: Screen mode; bottom row: Structured mode. Columns: Abundant, Moderate, and Sparse difficulty levels.}
  \label{fig:exploration-curves}
  \vspace{-13pt} 

\end{figure*}

Our primary results, summarized in Table~\ref{tab:avg_results_by_agent} and visualized in Figure~\ref{fig:exploration-curves}, report human-normalized UI Functional Discovery (\abrnormUniqueElements). The scores are normalized using the average human exploration performance (Table~\ref{tab:human_scores}), highlighting the significant variability between novice and expert human runs, establishing an expert-level performance benchmark of 114\% (Structured) and 117\% (Screen) for \abrnormUniqueElements. The absolute values for \abruniqueElements are reported in Table~\ref{tab:avg_results_by_agent_abs}., We also report \abruniqueFunctionalitiesTested at $2,000$ steps in Table~\ref{tab:avg_uft_by_agent} for the agents. The current implementation \abruniqueFunctionalitiesTested is based on the Structured mode action space, and would require further development to support the Screen mode. 

\begin{table}[h]
\centering
\setlength{\tabcolsep}{3pt} 
\caption{Human-normalized scores (\%) for UI Functional Discovery (\abrnormUniqueElements) at 2,000 steps for each level (Abundant(A), Moderate(M), Sparse(S)), and mean normalized scores across all levels at 500, 1,000, and 2,000 steps. Results for a single seed.
}

\label{tab:avg_results_by_agent}
\begin{tabular}{lccc|ccc}
\toprule
 & & & & \multicolumn{3}{c}{\textbf{Levels Average}} \\
\textbf{Agent} & \textbf{A@2k} & \textbf{M@2k} & \textbf{S@2k} & \textbf{@500} & \textbf{@1k} & \textbf{@2k} \\
\midrule
\multicolumn{7}{l}{\textbf{\textit{Structured Mode}}} \\
DFS        & 15.7 & 21.3 & 2.4 & 7.5 & 10.1 & 13.2 \\
random     & 44.1 & 17.4 & 3.7 & 14.1 & 18.1 & 21.7 \\
h-random* & 70.9 & 25.3 & 5.0 & 22.0 & 27.3 & 33.7 \\
BFS        & \textbf{89.5} & 30.5 & 5.8 & 27.2 & 38.6 & 41.9 \\
UIEx-AlGo & 80.4 & \textbf{74.9} & \textbf{76.4} & \textbf{49.5} & \textbf{60.8} & \textbf{77.2} \\
\midrule
\multicolumn{7}{l}{\textbf{\textit{Screen Mode}}} \\
random     & 37.6 & 9.0 & 3.1 & 10.4 & 13.5 & 16.6 \\
h-random* & 67.6 & 23.3 & 5.2 & 21.5 & 26.8 & 32.1 \\
GUI-Bee$\ddagger$*     & \textbf{69.9} & 25.8 & 5.3 & \textbf{22.9} & 28.1 & 33.7 \\
UIEx-AlGo & 59.3 & \textbf{60.1} & \textbf{57.6} & 21.0 & \textbf{37.3} & \textbf{59.0} \\
\bottomrule
\end{tabular}
\flushleft
{\footnotesize * agent has access to oracle information}
\vspace{-8pt} 
\end{table}


\begin{table}[h]
\centering
\setlength{\tabcolsep}{3pt}
\caption{UI Functional Discovery counts after 1 hour of exploration in \textit{Abundant} level for human participants. We report results for three levels of expertise: novice, intermediate, and expert, alongside the average used as base to normalize \abrnormUniqueElements scores.}
\label{tab:human_scores}
\begin{tabular}{lcccc}
\toprule
\textbf{Mode} & \textbf{Novice} & \textbf{Intermediate} & \textbf{Expert} & \textbf{Base (Avg)} \\
\midrule
Structured & 2196 & 2228 & 2722 & 2382 \\
Screen & 1458 & 1481 & 1880 & 1606 \\
\bottomrule
\end{tabular}
\end{table}

\begin{table}[h]
\centering
\setlength{\tabcolsep}{3pt} 
\caption{Scores for \uniqueFunctionalitiesTested (\abruniqueFunctionalitiesTested) at 2,000 steps for each level (Abundant(A), Moderate(M), Sparse(S)), and mean scores across all levels at 500, 1,000, and 2,000 steps. Results are reported for a single seed for agents in Structured mode and Screen mode for agents having access to perfect bounding box information.}
\label{tab:avg_uft_by_agent}
\begin{tabular}{lccc|ccc}
\toprule
& & & & \multicolumn{3}{c}{\textbf{Levels Average}} \\
\textbf{Agent} & \textbf{A@2k} & \textbf{M@2k} & \textbf{S@2k} & \textbf{@500} & \textbf{@1k} & \textbf{@2k} \\
\midrule
\multicolumn{7}{l}{\textbf{\textit{Structured Mode}}} \\
DFS        & 1.4 & 4.5 & 0.2 & 1.3 & 1.7 & 2.0 \\
random     & 8.8 & 5.6 & 2.7 & 2.4 & 3.6 & 5.7 \\
h-random* & 17.2 & 9.2 & 3.0 & 4.5 & 6.9 & 9.8 \\
BFS        & 16.8 & 7.7 & 1.7 & 5.4 & 8.0 & 8.7 \\
UIEx-AlGo & \textbf{34.2} & \textbf{26.4} & \textbf{29.6} & \textbf{13.8} & \textbf{19.4} & \textbf{30.1} \\
\bottomrule
\end{tabular}
\flushleft
{\footnotesize * agent has access to oracle information}
\vspace{-8pt} 
\end{table}

\ouralgo achieves the highest mean normalized scores across difficulty levels at 1,000 and 2,000 steps in both Structured and Screen modes, surpassing all baseline agents, even those leveraging oracle information. However, no single agent consistently outperforms the rest in all scenarios. \ouralgo, however does outperform the rest on \abruniqueFunctionalitiesTested, highlighting the fact that the algorithm interacts with different types of elements without repetition in the Structured mode.

In Abundant level, simple heuristic agents such as heuristic-random and BFS, outperform \ouralgo, particularly in the early stages of exploration (first 500 steps). This result suggests that abundant interactive elements allow heuristic strategies to rapidly discover functionalities. Specifically, BFS excels initially due to quick access to diverse functionalities directly reachable within a few levels of URL depth, plateauing only when delving into deeper content exploration (e.g., repository internals). Conversely, DFS performs poorly, becoming quickly trapped in repetitive depth interactions due to an early redirect to the vast help section of Gitlab.

\ouralgo demonstrates significant advantages in the Sparse level, where deliberate, multistep actions (e.g., creating users or projects) are essential to discover deeper functionalities. This indicates its ability to prioritize meaningful interactions effectively. \ouralgo achieves notably lower performance in the Screen mode compared to the Structured mode, underscoring the greater challenges posed by purely visual interaction and the need for improved visual grounding in agents.

Interestingly, the random selection baseline achieves relatively high scores in Abundant level (up to 44\% \abrnormUniqueElements), emphasizing that even naive strategies can produce substantial exploration in highly interactive environments. GUI-Bee$\ddagger$ remains slightly more effective than the heuristic-random approach, likely due to the complexity and the large action space of GitLab's interface. 

Overall, while \ouralgo consistently delivers strong exploration performance across modes and levels, the results reveal significant room for improvement (up to 22\% behind expert human exploration in Structured and 40\% in Screen modes). These results set a clear benchmark and motivate future advancements in UI exploration algorithms.



\subsection{Ablation \& analysis}
\label{sec:ablation}

\begin{figure}[t]
    \centering
    \includegraphics[height=4.4cm]{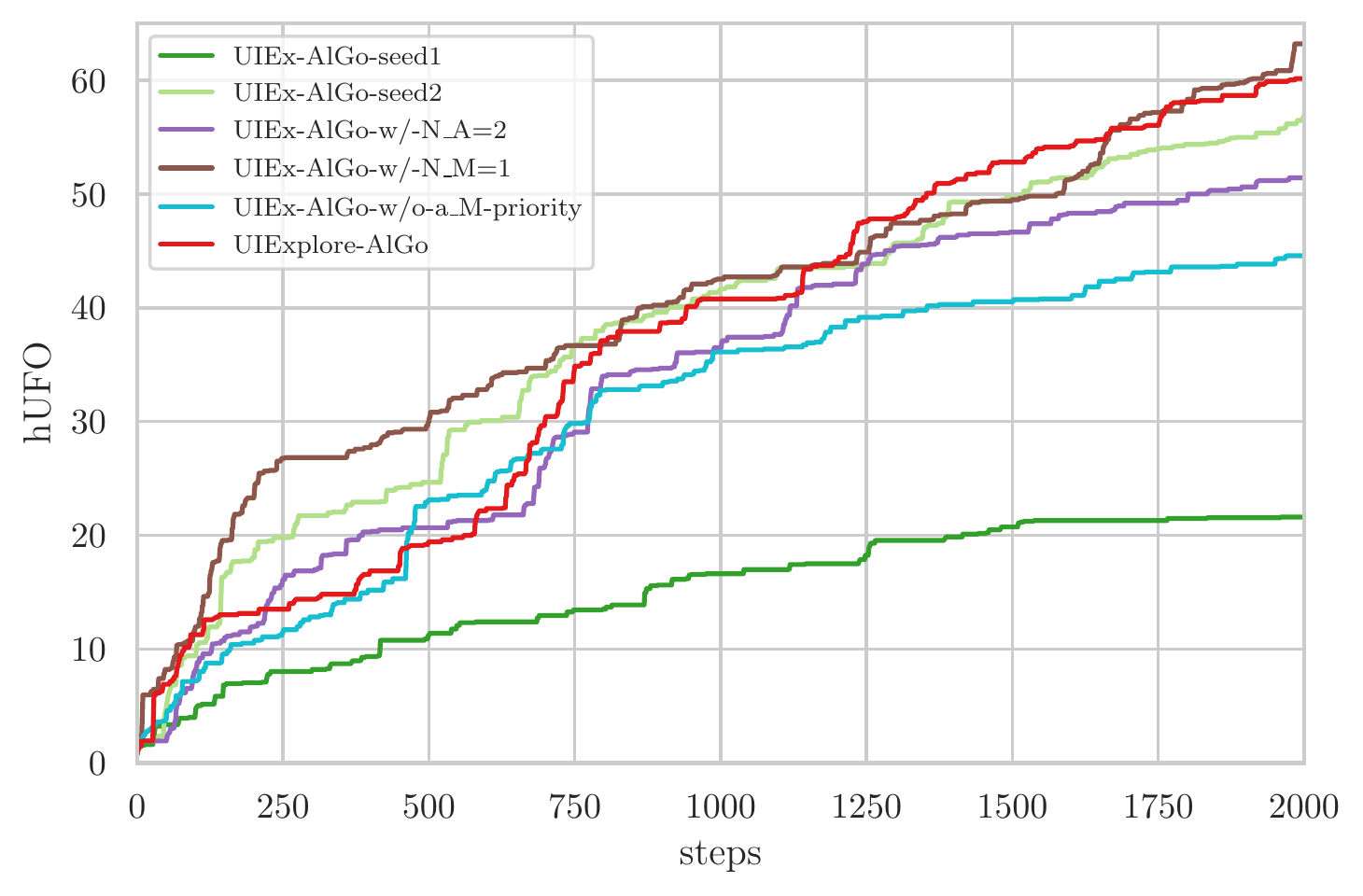}
    \vspace{-9pt} 
    \caption{Ablation results on the Screen–Moderate setting: human-normalized UI Functional Discovery (\uniqueElements) over 2,000 atomic actions. We compare the default UIEx-AlGo (solid green) against three ablations (i) single-macro horizon (\(N_M{=}1\)), (ii) tight atomic-action budget (\(N_A{=}2\)), and (iii) no priority/novelty, as well as two additional random seeds.}
    \vspace{-20pt} 
    \label{fig:correlation}
\end{figure}

To assess the robustness and design choices of \ouralgo, we performed three targeted ablations in the Screen–Moderate setting (Figure~\ref{fig:correlation}). We also report results over three random seeds to characterize variance.

\textbf{Seed Variability}.
Across three runs of UIEx-AlGo, two seeds achieved comparable performance (within $\pm5\%$ of the default), while Seed 1 underperformed, matching GUI-Bee$\ddagger$ and heuristic-random. Inspection revealed that Seed 1 failed to execute the \texttt{create\_project} macro in all three high-priority attempts (blank project, template import, continuation), due to novelty-based avoidance of retrying incomplete macros. This suggests that our current macro-completion feedback is insufficiently corrective, introducing high variance that lowers confidence in single-seed ablation interpretations. However, apart from Seed 1, all runs, including all ablations, successfully created a repository within the first 250 actions, indicating that early bootstrap success is a key driver of performance gap.

\textbf{Macro-Action Horizon (\(N_M=1\))}.
Setting \(N_M{=}1\) (one macro before selecting a new frontier) yields performance nearly on par with baseline heuristics, as expected: frequent frontier re-selection increases the chance of quickly discovering high-value UI segments. 

\textbf{Atomic-Action Budget (\(N_A=2\))}.
Capping atomic actions per macro to 2 produces a slight drop in overall coverage (–5\% at 2 k steps). Qualitative traces show that the agent generates partial macros, trying to continue the previous intent, requiring stitching tasks such as ``Create repository'' via multiple macros. With only two atomic steps, completion intent often spills over additional selections, reducing efficiency. However, the decline is modest, which implies that our prompting encourages continuation of the previous intent (Sect.~\ref{app:uiexplorer}).

\textbf{Priority \& Novelty Removal}.
 When both macro-action prioritization (LLM importance) and novelty scoring are ablated, selecting frontier and in-state macros uniformly at random, we observe a significant performance collapse (–10\% at 2 k steps). This confirms that prioritization is critical: without it, the agent drifts into redundant or low-value UI macro-actions.





\section{Call for applications}
\label{sec:applications}

We encourage the community to look beyond leaderboard scores and use \ourbench{} and its exploration dataset as a platform for broader research, including but not limited to:

\textbf{Downstream Exploitation with Shared Exploration Data.} Develop and compare algorithms that leverage our fixed exploration dataset for solving downstream tasks. For example, explore retrieval-augmented planning (RAG) for conditioning agents on prior exploratory experience to improve task-solving performance.

\textbf{Self-Supervised Metrics.} Design and evaluate new self-supervised metrics to measure exploration progress at test time, including metrics not reliant on privileged structural information.

\textbf{Reinforcement learning.} Train agents with RL to maximize \abruniqueElements and learn a better exploration policy, without handcrafted novelty heuristics.

\textbf{Intrinsic Reward Learning.} Train agents to maximize exploration-specific rewards (e.g., novelty, coverage, or “assumption discovery”). For instance, our dataset enables agents to analyze macro-action outcomes: if an agent predicts a button is for “merge” but the actual outcome differs, it can detect such mistaken assumptions and build intrinsic rewards for discovering surprises. We can train policies that seek out and correct the agents own misconceptions.

We welcome creative uses of \ourbench for new agent designs, metric development, or experimental protocols. All resources are publicly released to enable reproducibility and community-driven advancement.

\section{Discussion and Future Work}
\label{sec:discussion}

We introduced \ourbench{}, a dedicated benchmark for evaluating user interface exploration by autonomous agents. Motivated by a growing need in the community, \ourbench{} provides a standardized environment for isolating and studying exploration behaviors. We launch a suite of baselines alongside our proposed algorithm \ouralgo{}, to highlight key challenges and establish reference points for progress. \ouralgo{} combines novelty- and preference-based prioritization with a hierarchical approach to efficient, structured exploration.

Our current setup focuses on a single application (GitLab), which, while realistic, may not fully capture the diversity of UI exploration challenges. Moreover, our algorithm does not yet incorporate policy-based backtracking, which could further enhance performance.

Future extensions of \ourbench could include additional applications and introduce more subtle or less conventional interactive elements to probe agent generalization and robustness. 

We believe that our work aligns with recent calls to move beyond static datasets and toward continuous, experience-driven learning~\cite{silver2025welcome}, which holds promise for agents capable of long-term adaptation and discovery.

We invite the community to contribute with new environments, metrics, and agents to \ourbench, and to consider our call for action Section \ref{sec:applications}. Strengthening UI exploration is a key step toward building truly autonomous and robust agents for real-world digital tasks.

\newpage
\clearpage

\section*{Software, Data and Leaderboard}

\textbf{Benchmark Release.} We will release the full \ourbench{} benchmark environments, our exploration data set, open source tools to load and explore the collected dataset and the reporting code, within two months of acceptance.
\textbf{Public Leaderboard.} A continuously updated leaderboard will be hosted, ranking submissions by \uniqueElements{} at 2,000 steps, with additional reporting required at 500 and 1,000 steps. Scores must be provided separately for the DOM and GUI modes, and each submission must include a link to the trajectory data and, preferably, the code.

\section*{Limitations}

\textbf{Budget Constraints.} Due to the limitations of the compute and API rate, our experimental evaluation was limited in the number of seeds, agent variants and ablation runs. For efficiency, we used a hybrid setup: GPT-4o for state description and macro-action generation, and Claude for action selection, enabling balanced querying under usage caps. Although this configuration was fixed across runs, it would be valuable future work to systematically evaluate different combinations of models for each component to better understand their respective contributions.

\textbf{Scope.} The benchmark currently targets a single application (GitLab) and supports exploration with goto commands. Although this reflects real-world usage, it may reduce exploration challenge compared to stricter navigation constraints. We plan to release a goto-free version to support more rigorous future comparisons.

\section*{Impact Statement}

\textbf{Positive Impact.} This work aims to advance the study of structured exploration in user interfaces, enabling agents to autonomously build actionable knowledge from real-world environments. Potential applications include more reliable task automation, better user accessibility testing, and scalable collection of experience data for training downstream models.

\textbf{Potential Risks.} As with any tool that improves understanding of the environment, there is a risk of misuse such as aggressive scraping and automating spam-like interactions. However, our benchmark is designed for sandbox environments and focuses on measuring functional discovery rather than real-world automation. We encourage future work to build safeguards when deploying such agents broadly.

\section*{Acknowledgments}
We thank Surbhi Kamboj and Tomasz Religa for their assistance in collecting the exploration data.

\bibliography{main}
\bibliographystyle{icml2025}

\newpage
\appendix
\onecolumn





\section{Extended results}
\label{app:extended-results}

The absolute \uniqueElements counts can be found in Table~\ref{tab:avg_results_by_agent_abs}.

\begin{table}[t]
\centering
\setlength{\tabcolsep}{3pt} 
\caption{Scores for \uniqueElements (\abruniqueElements) at 2,000 steps for each level (Abundant, Moderate, Sparse), and mean scores across all levels at 500, 1,000, and 2,000 steps. Results are reported for a single seed.}
\label{tab:avg_results_by_agent_abs}
\begin{tabular}{lccc|ccc}
\toprule
& & & & \multicolumn{3}{c}{\textbf{Levels Average}} \\
\textbf{Agent} & \textbf{A@2k} & \textbf{M@2k} & \textbf{S@2k} & \textbf{@500} & \textbf{@1k} & \textbf{@2k} \\
\midrule
\multicolumn{7}{l}{\textbf{\textit{Structured Mode}}} \\
DFS        & 375.0 & 508.0 & 57.0 & 179.3 & 241.0 & 313.3 \\
random     & 1050.0 & 414.0 & 88.0 & 335.7 & 431.3 & 517.3 \\
BFS        & 2132.0 & 726.0 & 137.0 & 648.3 & 919.7 & 998.3 \\
heuristic-random & 1689.0 & 602.0 & 118.0 & 523.3 & 651.3 & 803.0 \\
UIExplore-AlGo & 1914.0 & 1784.0 & 1821.0 & 1180.0 & 1447.7 & 1839.7 \\
\midrule
\multicolumn{7}{l}{\textbf{\textit{Screen Mode}}} \\
random     & 604.0 & 144.0 & 50.0 & 167.0 & 217.3 & 266.0 \\
heuristic-random & 1086.0 & 375.0 & 84.0 & 346.0 & 430.3 & 515.0 \\
GUI-Bee$\ddagger$     & 1123.0 & 415.0 & 85.0 & 368.3 & 450.7 & 541.0 \\
UIExplore-AlGo & 952.0 & 966.0 & 925.0 & 337.3 & 598.7 & 947.7 \\
\bottomrule
\end{tabular}
\end{table}

\section{Human evaluation}
\label{app:human-evaluation}

We collected exploration trajectories from three human participants with different levels of expertise: novice, intermediate, and expert. Participants were asked to ``explore the Gitlab website to discover the functionalities exposed by the platform'' in Abundant level. The data were collected for approximately one hour. Observations were captured at a rate of at most one observation per second. The observation captures were paused while the web pages were being loaded to avoid redundant data captures.

\begin{figure}[t]
    \centering
    \fbox{
        \includegraphics[width=0.45\linewidth]{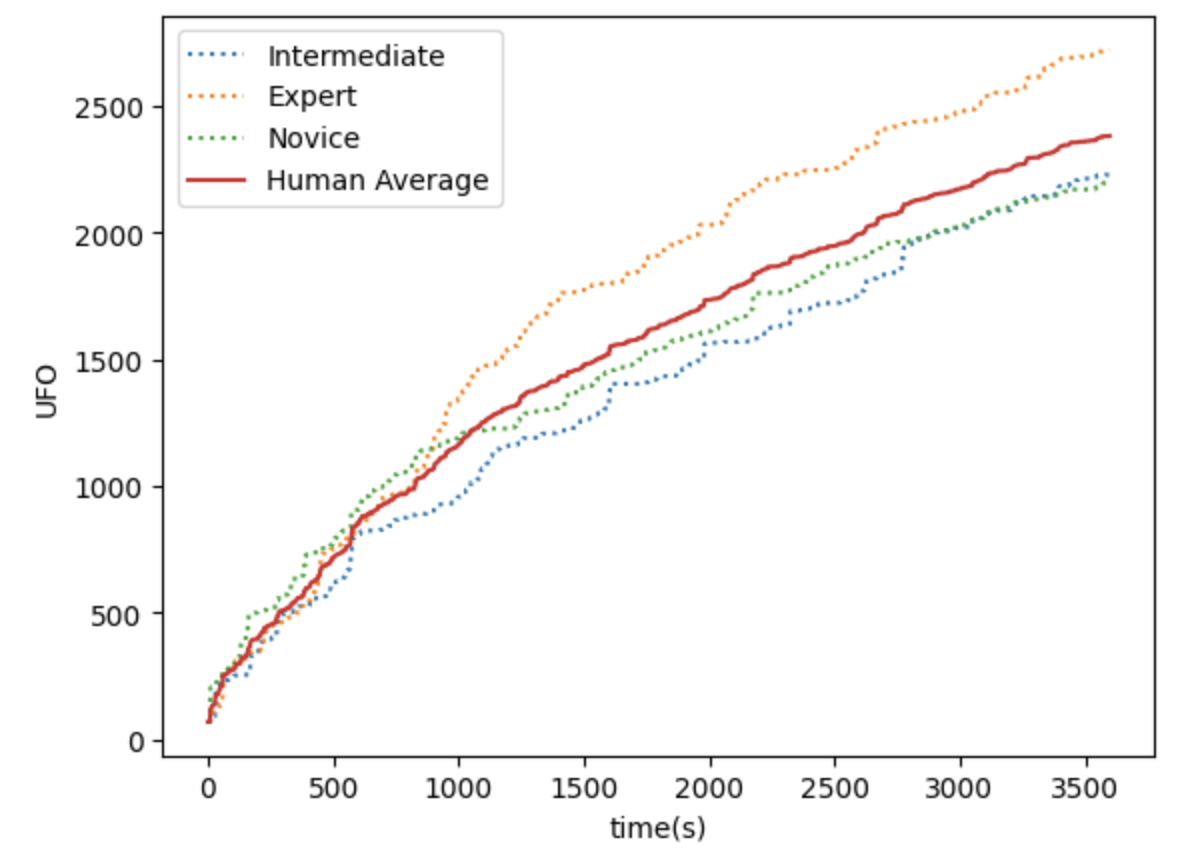}
    }
    \fbox{
        \includegraphics[width=0.45\linewidth]{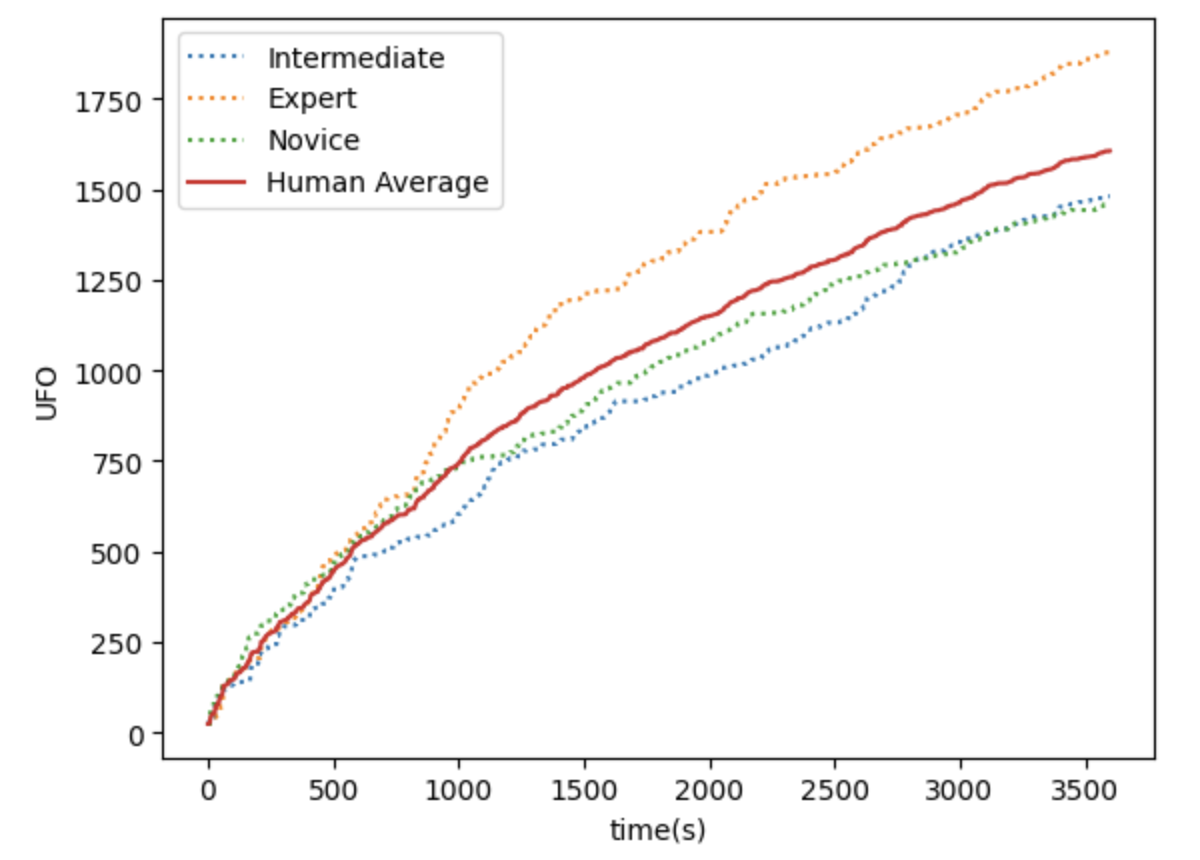}
    }
    \caption{\abruniqueElements by time in Structured mode (left) and Screen mode (right) for three human participants having various levels of expertise}
    \label{fig:human-ufo-by-time}
\end{figure}

The graph of human exploration over 1 hour can be found in Fig.~\ref{fig:human-ufo-by-time}. Since the different human annotators captured a slightly different number of observations, the x-axis has been renormalized to be 1 hour.

\section{Metric Formalism}
\label{app:metric-formalism}

Our primary metric, UFO, is based on functionalities exposed by an application in a given observation. A functionality is defined as a group of individual actions that share transitional and visual characteristics.

\begin{figure}[t]
    \centering
    \fbox{
        \includegraphics[width=0.45\linewidth]{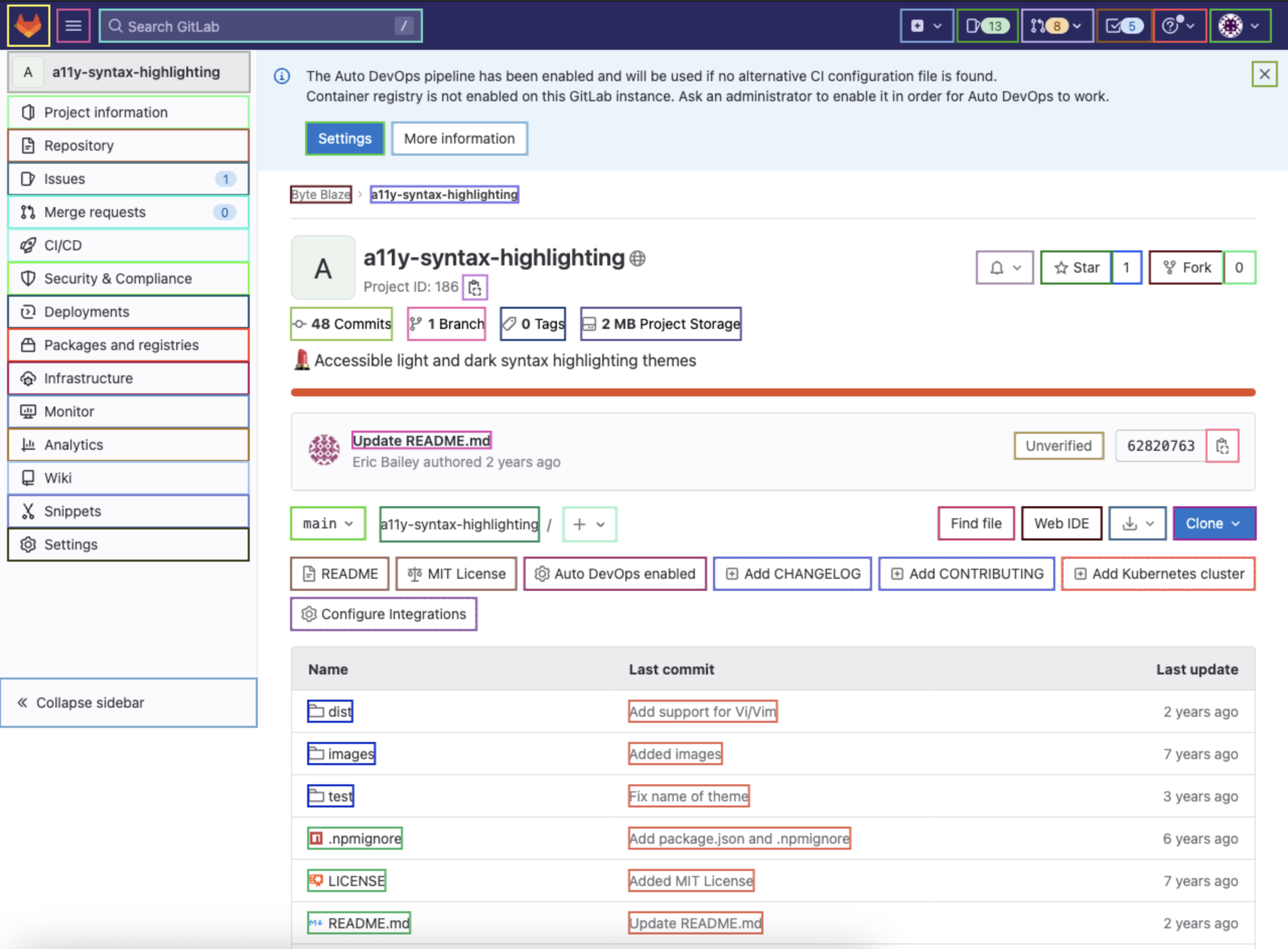}
    }
    \fbox{
        \includegraphics[width=0.45\linewidth]{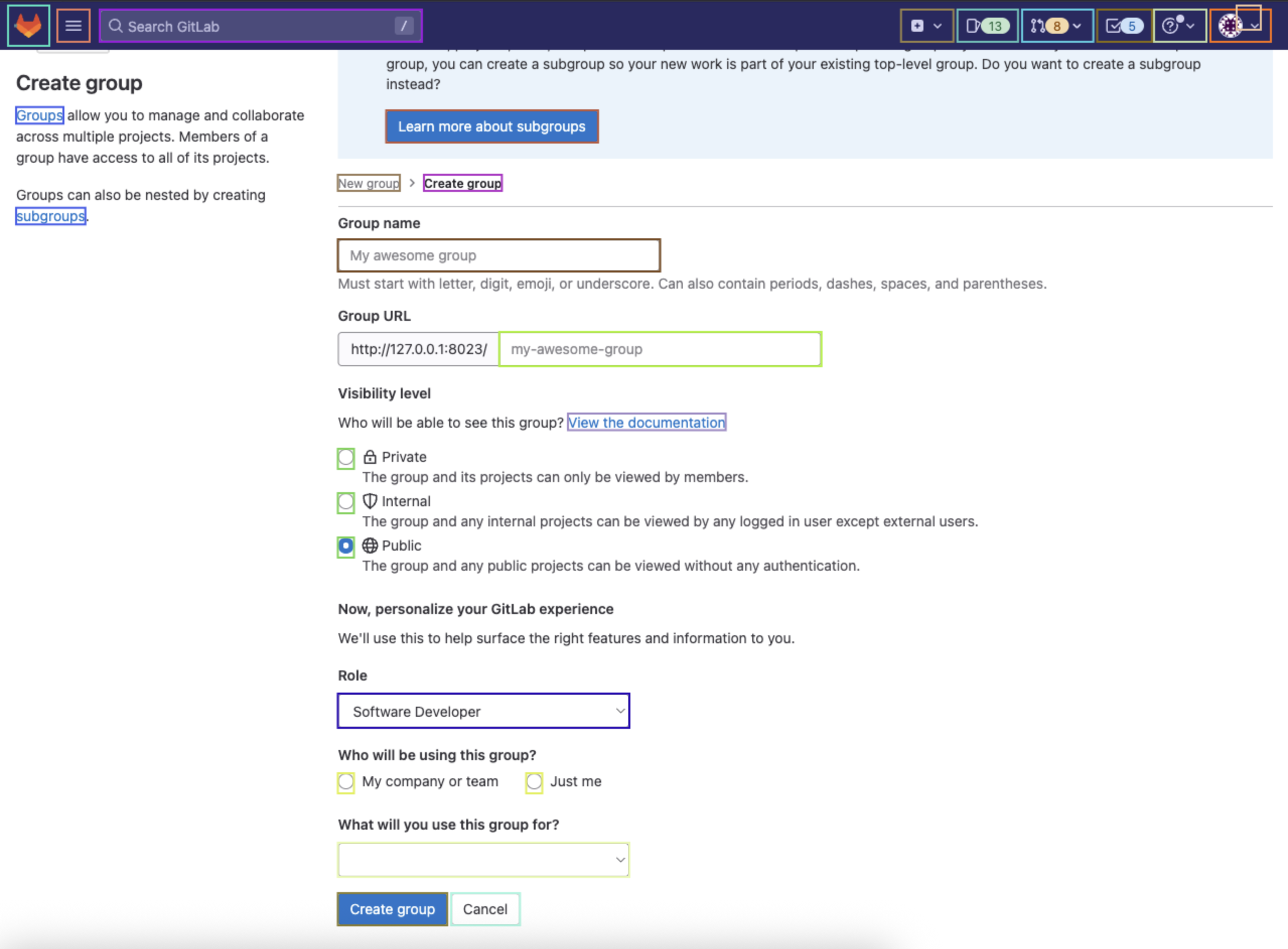}
    }
    \caption{Functionality groupings in project page (left) and create group page (right). Within each image, boxes of the same color are grouped under the same functionality.}
    \label{fig:metric-groupings}
\end{figure}

For Gitlab environment, the transitional characteristics are extracted using DOM tags. All \texttt{button}, \texttt{input}, \texttt{select}, and \texttt{textarea} tags that are not disabled are considered as providing inputs for web state transitions. The \texttt{a} (link) tags are grouped based on URL-patterns specific to Gitlab. For example, links to \texttt{org1/project1} and \texttt{org2/project2} are treated to provide the same functionality -- navigation to a project, \texttt{\{org\}/\{project\}}. The visual characteristics are extracted primarily by observing the class attribute associated with the DOM tags. However, this is insufficient for addressing complex styling within Gitlab. To address the corner cases, hand-crafted rules are applied for grouping visual characteristics based on DOM attributes. Additional examples of groupings on the Gitlab ``project page'' and ``create group'' can be found in Fig.~\ref{fig:metric-groupings}.

\begin{figure}[t]
    \centering
    \fbox{
        \includegraphics[width=0.45\linewidth]{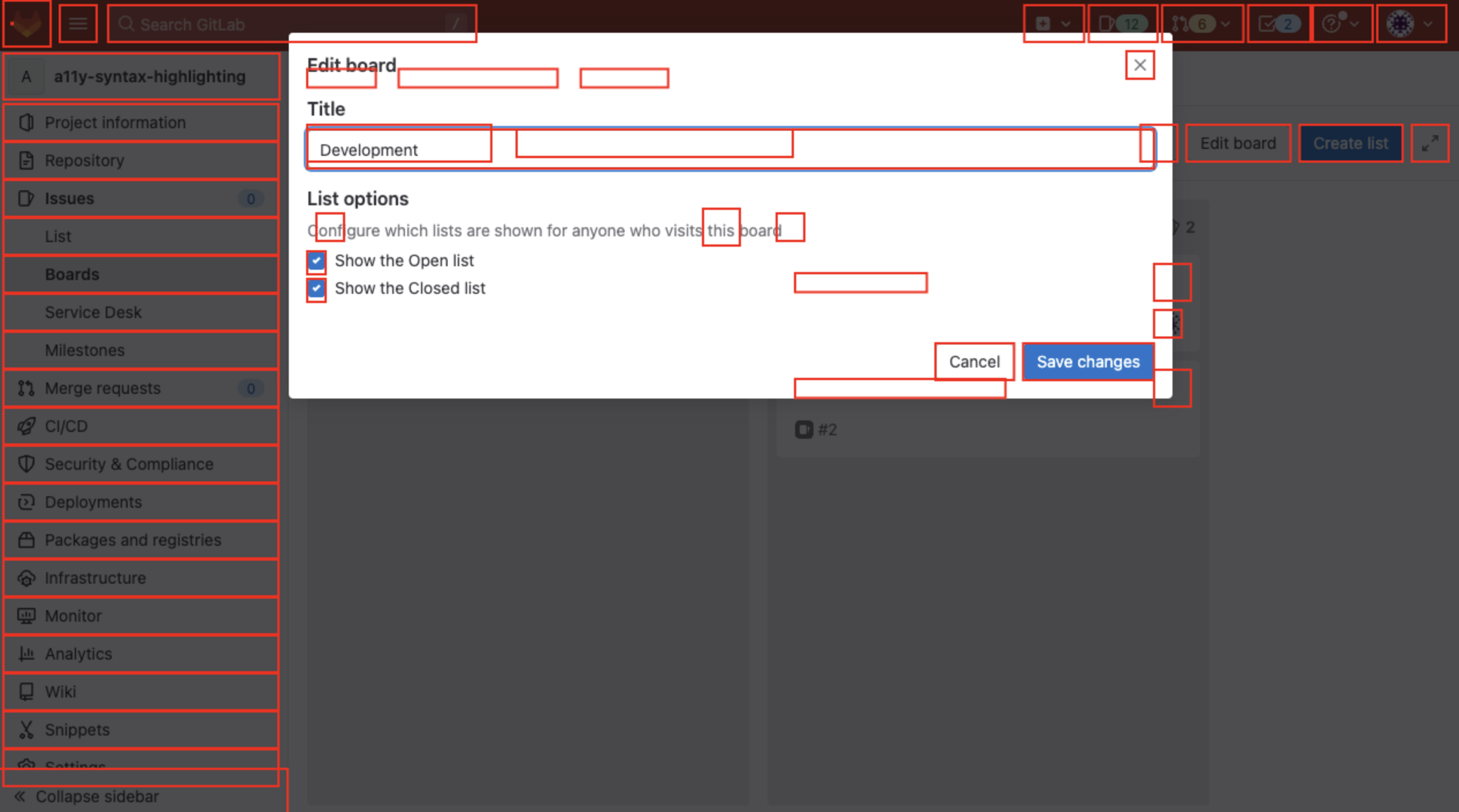}
    }
    \fbox{
        \includegraphics[width=0.45\linewidth]{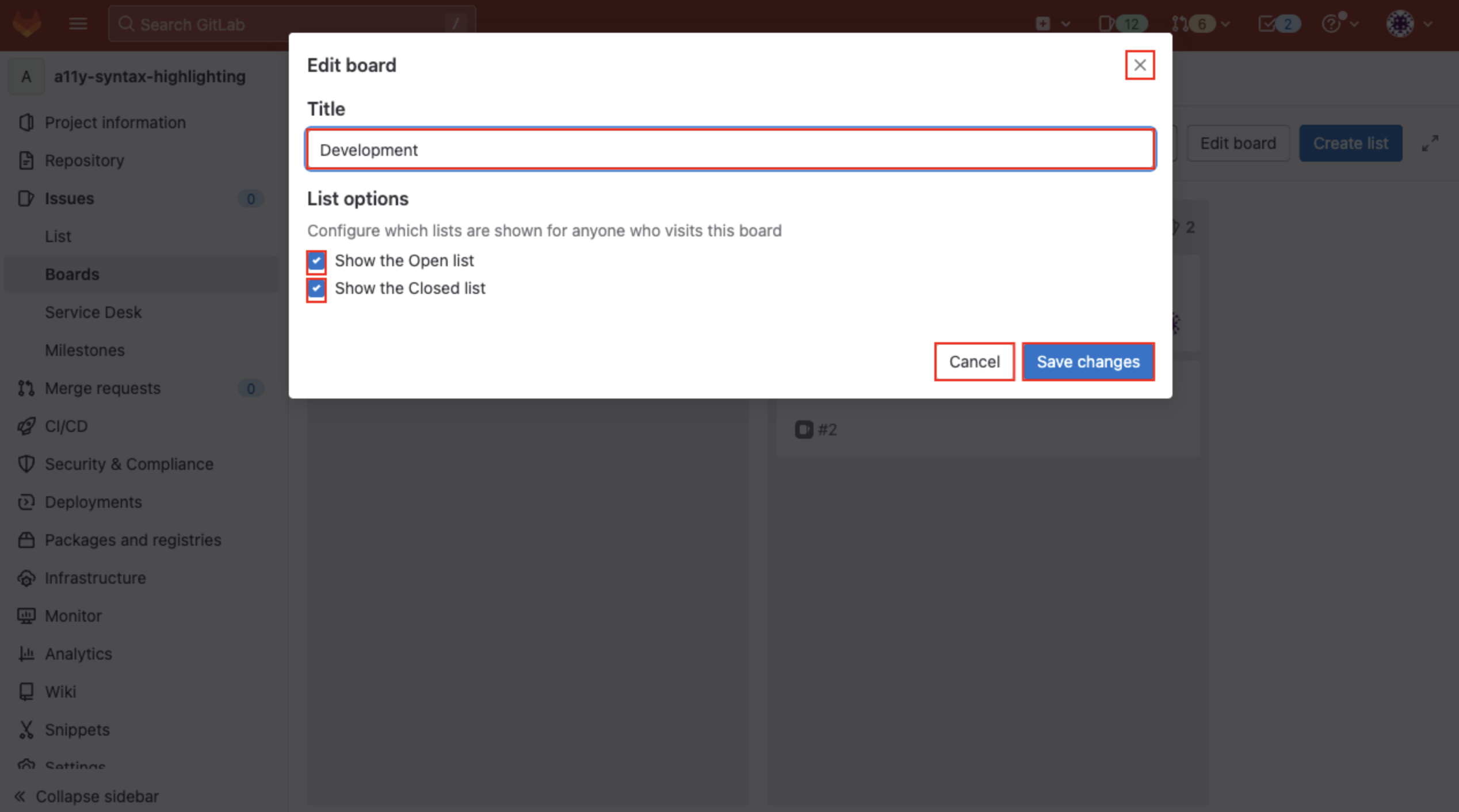}
    }
    \caption{UFO Structured Mode (left) includes all functionalities present in the DOM regardless of their visibility and interactivity status and UFO Screen Mode (right) includes functionalities that are visible and interactive}
    \label{fig:metric-formalism}
\end{figure}

Our metrics can be applied for both the Structured mode and the Screen mode. The Structured mode considers all the functionalities that are present in the DOM structure of the webpage. Screen mode only includes functionalities that are visible and actionable within the browser's viewport. An example showing this difference can be found in Fig.~\ref{fig:metric-formalism}.

\section{Environment Details}
\label{app:env-details}
\ourbench uses a customized version of the Webarena \cite{zhou2023webarena} Gitlab environment.

\begin{figure}[t]
    \centering
    \fbox{
        \includegraphics[width=0.45\linewidth]{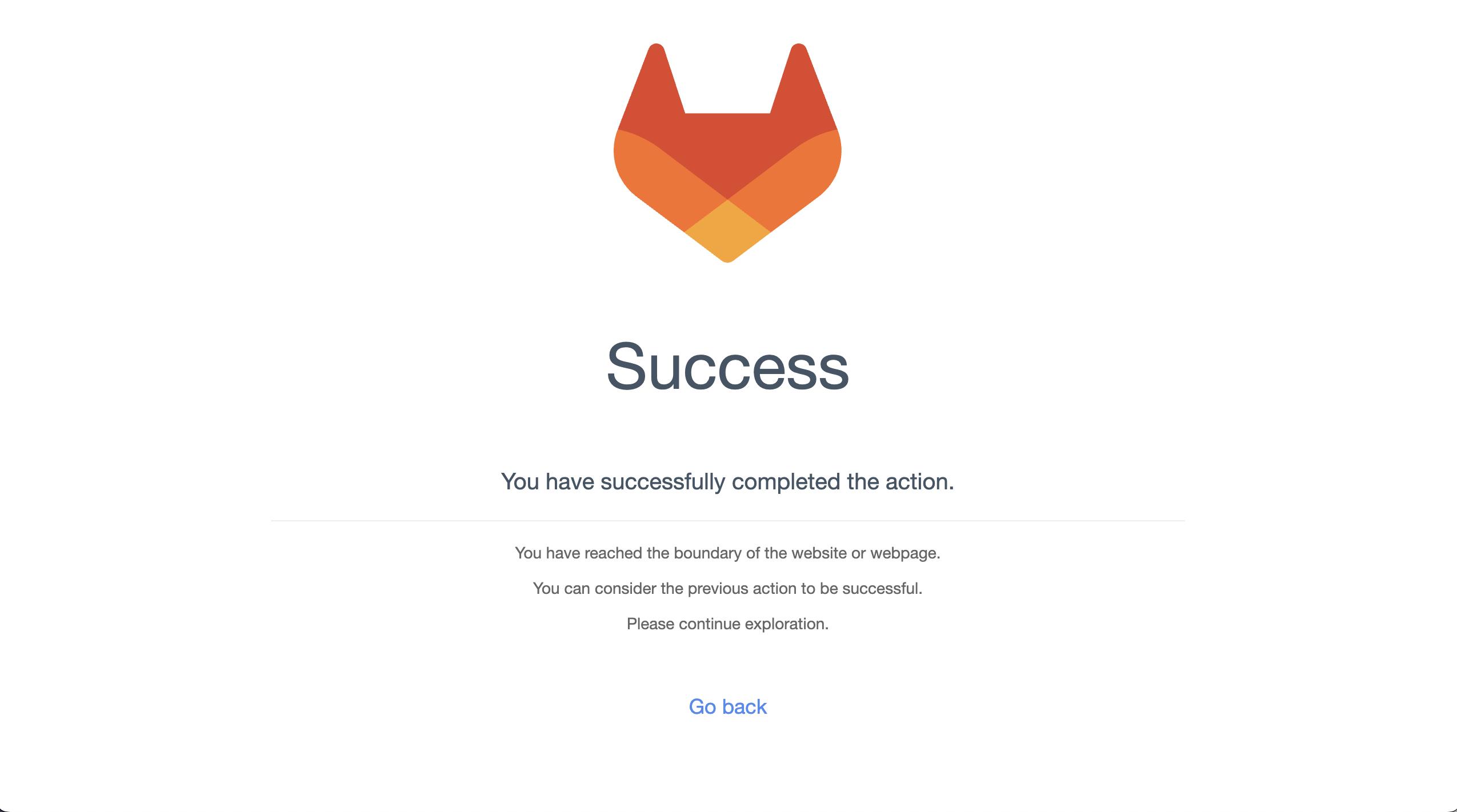}
    }
    \fbox{
        \includegraphics[width=0.45\linewidth]{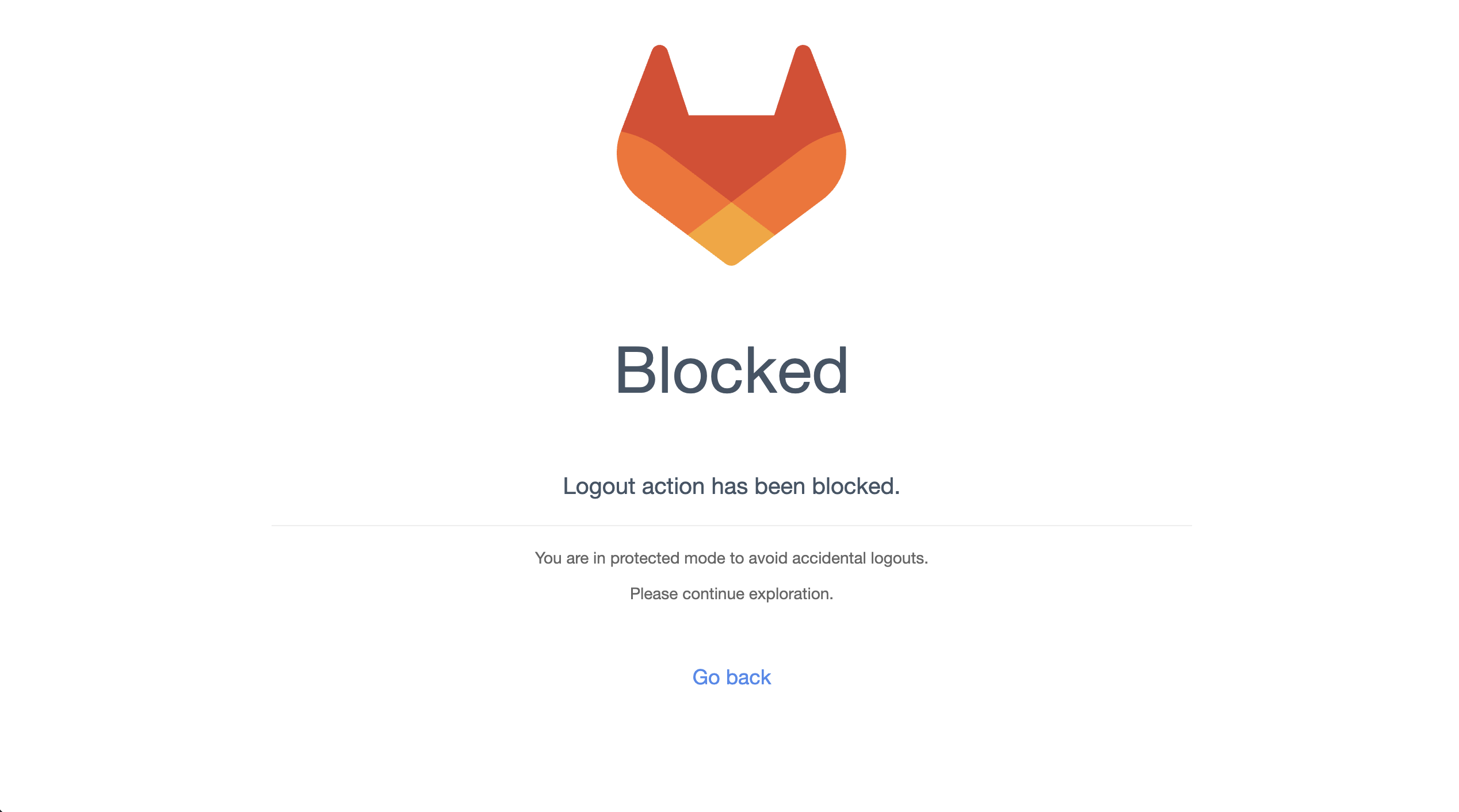}
    }
    \caption{Boundary Page (left) and Logout-disabled Page (right)}
    \label{fig:boundary-page}
\end{figure}

In order to contain agents within the environment, we redirect all access to external websites and file-upload dialog boxes to a boundary page. The boundary page displays a message stating that the action is to be considered successful, along with a home button and a go-back button. Additionally, in the Abundant and Moderate modes, all attempts to log-out are disabled and redirected to a log-out-disabled page. Fig.~\ref{fig:boundary-page} shows the visual representation of these two pages.

\textbf{Environment reset details.}

\begin{enumerate}
  \item Abundant mode resets to the default Webarena's Gitlab instance.
  \item Moderate mode resets to the default Webarena's Gitlab instance followed by the deletion of all Gitlab projects. All users are retained.
  \item Sparse mode resets to the default Webarena's Gitlab instance followed by the deletion of all Gitlab projects and users.
\end{enumerate}


\section{D3C Discussion}
\label{app:d3c-analysis}

\begin{figure}[t]
    \fbox{
        \includegraphics[width=0.45\linewidth]{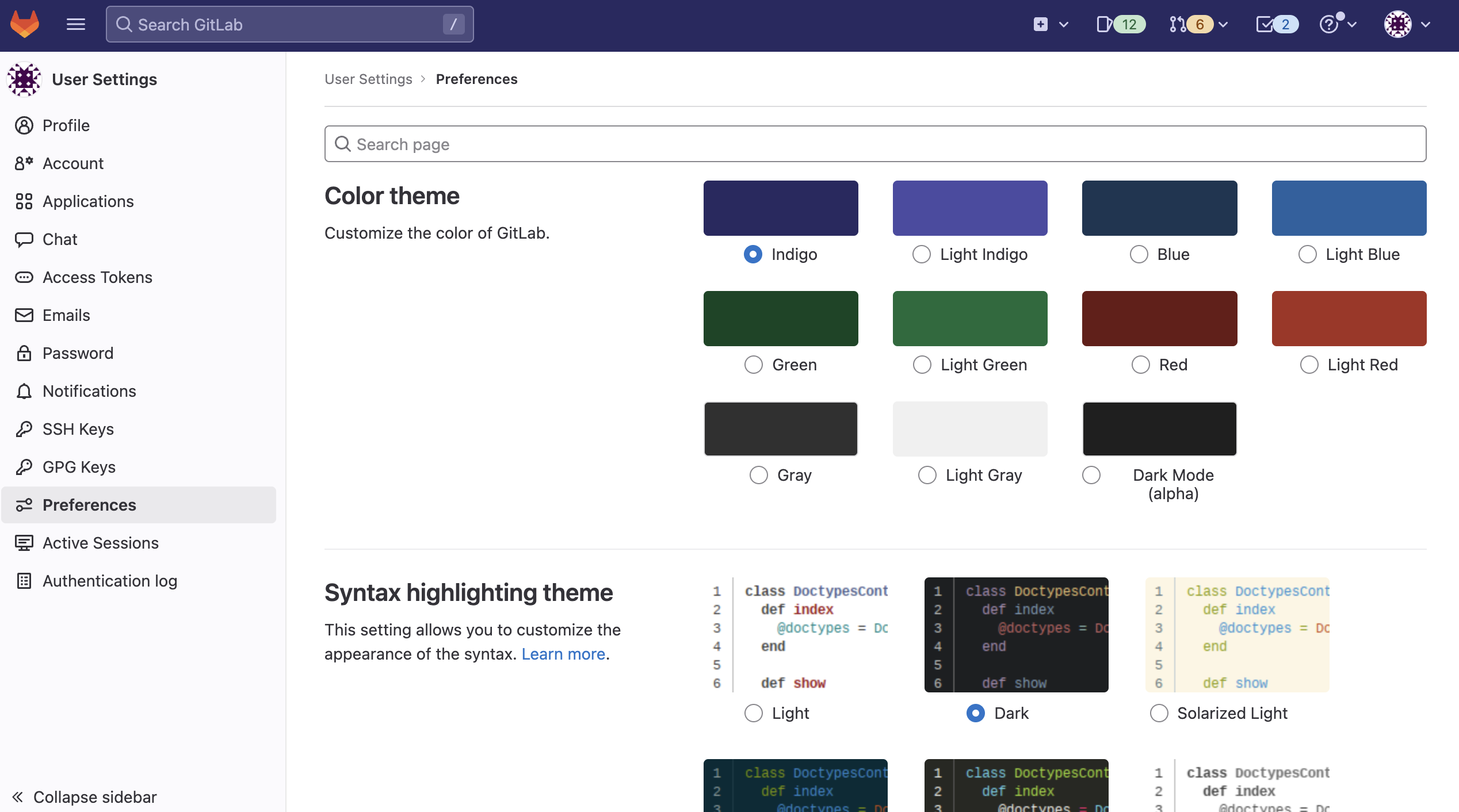}
    }
    \fbox{
        \includegraphics[width=0.45\linewidth]{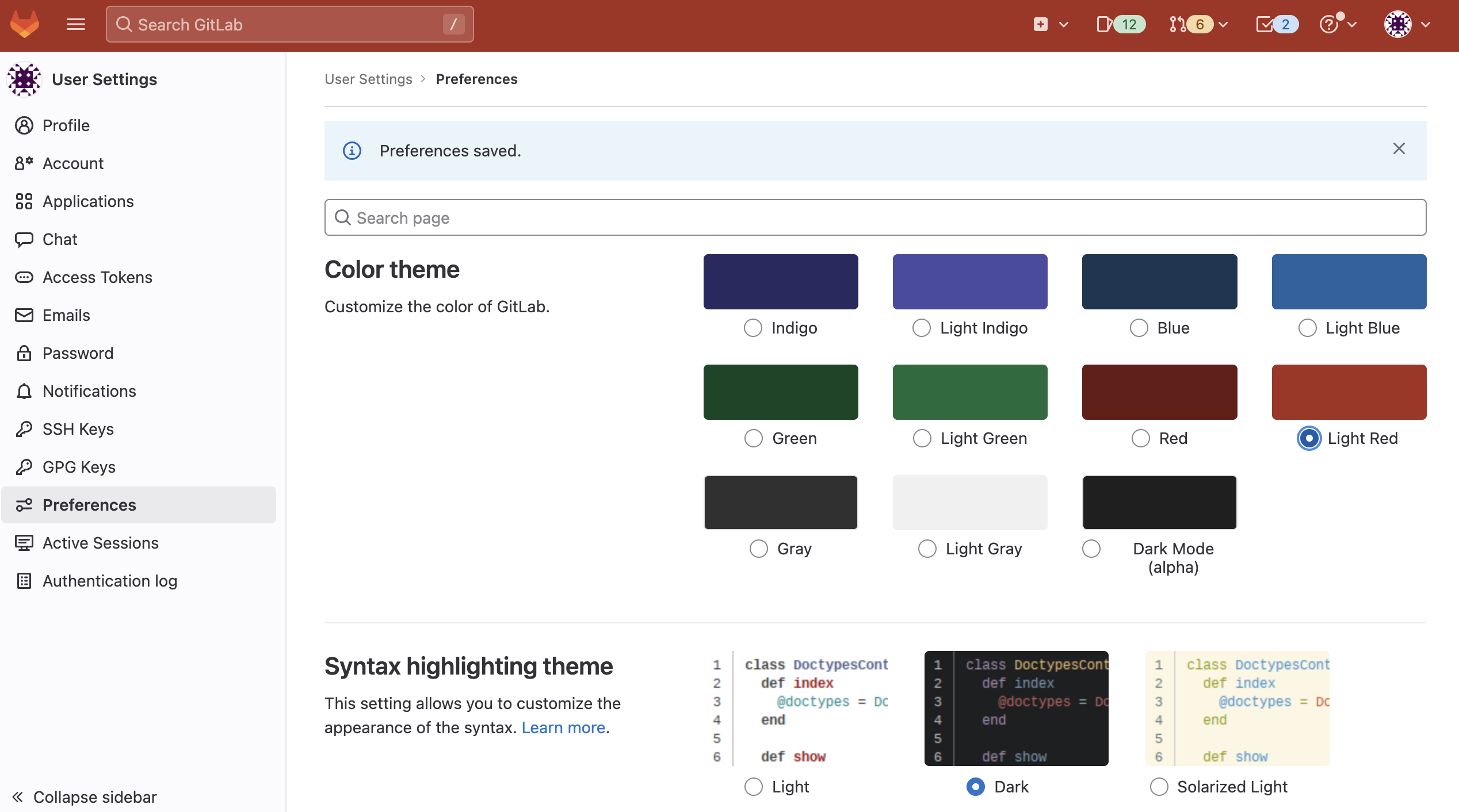}
    }
    \caption{Indigo theme (left) and Light Red theme (right) updates the DOM at \texttt{body} tag (depth 1)}
    \label{fig:d3c-analysis}
\end{figure}

Fig.~\ref{fig:d3c-analysis} shows an example in which changes in the user preference of \texttt{theme} affects D3C computation. Updating the \texttt{theme} modifies the \texttt{class} attribute of the \texttt{body} tag. Since this change occurs at depth 1, all pages visited under one \texttt{theme} is treated to be distinct from the same pages visited under another \texttt{theme}, thereby incentivizing the exploration agent to change themes often.

\section{UIExplorer Agent Details}
\label{app:uiexplorer}

Our proposed algorithm, \textsc{UIExplore-AlGo}, is designed for efficient and structured UI exploration, combining insights from hierarchical exploration strategies and novelty-driven search inspired by Go-Explore \cite{ecoffet2019go}. Its core features include macro-actions for high-level interaction, novelty-based prioritization, and systematic frontier state selection. We provide the pseudocode in Algorithm \ref{alg:uiexplorer}, and further detail its primary components and computation methodologies here.

\textbf{Core Exploration Loop.}
The algorithm operates by repeatedly describing the current UI state using GPT-4o to generate textual descriptions of states and macro-actions (higher-level UI interaction tasks). These macro-actions encapsulate sequences of atomic UI interactions (e.g., filling a form, creating a repository). The agent alternates between exploring novel macro-actions from the current state and periodically revisiting frontier states identified as promising based on their exploration potential.

\textbf{Knowledge Graph and Dataset.}
All states, macro-actions, and their resulting state transitions are recorded in a semi-structured dataset $G$, organized as a knowledge graph. We publicly release this dataset as part of our benchmark to facilitate downstream research, enabling agents to utilize past exploration experiences.

\textbf{Novelty-Based Macro-Action Selection.}
Novelty-driven macro-action selection involves three primary considerations:

\paragraph{- Predictive State Novelty} GPT-4o generates an anticipated future state description $s_f$ for each candidate macro-action. Let $\mathcal{F}$ represent embeddings of predicted future state descriptions and $\mathcal{R}$ embeddings of actually experienced state descriptions. We compute novelty via cosine similarity:

$$
\text{Novelty}_{state}(a) = 1 - \max_{r \in \mathcal{R}} \frac{f_a \cdot r}{\|f_a\| \|r\|}, \quad \text{where } f_a \in \mathcal{F}
$$

\paragraph{- State-Action Pair Novelty} We calculate novelty between candidate state-action embeddings $\mathcal{C}$ and embeddings of previously executed state-action pairs $\mathcal{P}$:

$$
\text{Novelty}_{action}(a) = 1 - \max_{p \in \mathcal{P}} \frac{c_a \cdot p}{\|c_a\| \|p\|}, \quad \text{where } c_a \in \mathcal{C}
$$

\paragraph{- Importance Ranking by GPT-4o} GPT-4o also ranks macro-actions based on their predicted complexity and exploration depth potential. The top-ranked actions by GPT-4o are given additional priority to balance purely novelty-driven selection.

\paragraph{Macro-action Choice}
Macro-actions are selected in two stages: a novelty-based filtering step followed by GPT-based prioritization.

\textbf{- Filtering.}  
\begin{itemize}
    \item In the \textit{frontier-based setting}, we select the top 20 macro-actions with the highest novelty scores.  
    \item In the \textit{local exploration setting}, we select the top 30\% most novel macro-actions available at the current state.
\end{itemize}

\textbf{- Prioritization.}  
From the filtered set $\mathcal{A}_{\text{novel}} \subseteq \mathcal{A}$, the final macro-action is chosen based on GPT-4o’s ranking:

$$
a_{M}^{*} = \underset{a \in \mathcal{A}_{\text{novel}}}{\arg\max} \ \text{Rank}_{GPT}(a)^{-1}
$$

\noindent That is, we select the most important macro-action (according to GPT-4o) among the most novel ones.

\textbf{Frontier State Selection}
Periodically, the algorithm selects frontier states—previously encountered states judged promising based on novelty and potential exploration gains. We reuse the novelty computation described above, applied to all recorded states and their available macro-actions, choosing frontier states-actions with maximal exploration potential.

\textbf{Macro-action Execution and Atomic Actions}
The execution of macro-actions is handled by a secondary agent, Claude, optimized empirically for the direct execution of low-level UI actions. Claude's responsibility is executing sequences of atomic actions until the macro-action goal is completed or a step limit $N_A$ is reached.

\textbf{Backtracking and Goal-Conditioned Navigation}
To efficiently revisit frontier states, we leverage a built-in \texttt{goto} function. For applications lacking this capability, future extensions could incorporate learned goal-conditioned navigation policies as suggested in related work \cite{ecoffet2021first}.

\textbf{Implementation and Hyperparameters}
Full algorithm details, including prompt templates and additional hyperparameter settings, are available publicly alongside the benchmark release to facilitate reproducibility and extension.

\subsection{Prompts}

\textbf{Describe state and macro-actions prompt:}

\begin{tcolorbox}[colback=gray!5!white, colframe=gray!75!black, boxrule=0.4pt, breakable]
\scriptsize\ttfamily

<system\_prompt>

You are an autonomous UI explorer. Your goal is to map out an application’s full feature set by systematically 
driving it from its current state into new states that unlock yet more functionality.

Guiding principles:

1. **Depth-first discovery**  
   Order your actions by their likelihood of leading you deeper into the app—unlocking new screens, flows, or features.  
   
2. **Complex over trivial**  
   Prefer composite tasks (e.g. “submit registration form”) over atomic ones (“click this icon”).  
   
3. **Cover all unique interactive types**  
   Ensure that each type of interactive element (e.g. buttons, links, icons, inputs) is exercised by at least one action. You don’t need a separate task for every instance—group similar elements into a single representative action (e.g. “click one repository link” or “try clicking a like button”). If needed, use actions like “click one of the X buttons” or “click each X button” to generalize.
   Even if you don’t know what an icon or button does, include an action to “interact with” it.  
   
4. **Keep it in-app**  
   Actions that navigate outside the app (external links, file uploads, downloads) go at the end.  
   
5. **ASCII only**  
   Use plain ASCII for all descriptions and actions.  
   
6. **Fallback on dead-ends**  
   If you hit a missing page or error boundary, issue a “go back” action.

When you receive a state you’ll reply in JSON:

```json
{
  "state": "<text description of the current screen, its inputs, icons, buttons and links>",
  "actions": [
    "<highest-priority action>",
    "... next action ...",
    "... lowest-priority action ..."
  ]
}
```

Your goal is to ensure all unique interactive element types are covered by at least one action, with grouping used when multiple similar elements are present.

</system\_prompt>

<user\_prompt>

Current application state

{{ state\_description }}

Last action taken
{{ last\_action }}

Generate your JSON response based on the system rules above. Focus first on sequences that open up new app modules (e.g. “complete registration”), then on other in-app features, then on external links.

</user\_prompt>

\end{tcolorbox}

\textbf{Act-agent prompt for Structured Mode:}

\begin{tcolorbox}[colback=gray!5!white, colframe=gray!75!black, boxrule=0.4pt, breakable]
\scriptsize\ttfamily

<system\_prompt>

You are **AxBrowser‑Agent**, a deterministic web‑automation specialist.
You only see the page through an **AXTree** snapshot that is refreshed
after every action.

TASK

1. Fulfil the user’s goal.

2. If the goal seems reached, finish with
   `<action>send\_msg\_to\_user("completed")</action>`
   
3. If you are stuck (e.g. the last 2‑3 actions had no effect, or the
   required element is missing) finish with
   `<action>report\_infeasible()</action>`

ACTION FORMAT

* Use the element number id (without the 'browsergym\_id\_' prefix) *
Return **exactly** the following XML tags per reply, nothing else:

```xml
<previous\_action\_outcome>...</previous\_action\_outcome>

<action\_description>...</action\_description>

<action>click("element\_id")</action>
```

Action space is provided below:

{{ action\_space }}

REASONING GUIDELINES (keep your chain‑of‑thought PRIVATE)

Always inspect the new AXTree (**Current page**)  after every action.

Detect dropdowns/menus:
If the clicked element now has expanded="true" or you see
new list / menu items, assume a sub‑menu opened.

Prefer selectors that match AX name, label, value, or role.
When several candidates exist, choose the one closest in the tree
to the element you just interacted with.

Avoid loops: if the URL AND top‑level AXTree snippet have not
changed after two actions, rethink or quit.

Stop if the **Current page** content reflects the ambiguous goal, otherwise, continue exploring.

Tasks can be ambiguous or slightly wrong; interpret generously.

</system\_prompt>

<user\_prompt>

\# Mini‑task

{{ goal\_object }}

Recent history (last steps, from oldest to newest. - n steps before.)

<action\_history>

$for h in action_history[::-1]$

  <step {{ h.idx }}>
  
    acted= <previous\_action{{ h.idx }}> {{ h.action }} </previous\_action{{ h.idx }}>

    saw  = <previous\_observation{{ h.idx }}> {{ h.observation }} </previous\_observation{{ h.idx }}>

  </step>

</action\_history>

Current page

URL: {{ open\_tabs }}

{{ state\_description }}

**Reply with the following XML tags, according to the instructions and action space:**

```xml

<previous\_action\_outcome>...</previous\_action\_outcome>

<action\_description>...</action\_description>

<action>...</action>

```

</user\_prompt>

\end{tcolorbox}

\textbf{Act-agent prompt for Screen Mode:}

\begin{tcolorbox}[colback=gray!5!white, colframe=gray!75!black, boxrule=0.4pt, breakable]
\scriptsize\ttfamily

<system\_prompt>

You are **GUIBrowser‑Agent**, a deterministic web‑automation specialist.
You only see the page through a **screenshot**.
You are a computer use agent.

TASK

1. Fulfil the user’s goal.

2. If the goal seems reached, finish with
   `<action>send\_msg\_to\_user("completed")</action>`
   
3. If you are stuck (e.g. the last 2‑3 actions had no effect, or the
   required element is missing) finish with
   `<action>report\_infeasible()</action>`

ACTION FORMAT

Return **exactly** the following XML tags per reply, nothing else:

```xml
<previous\_action\_outcome>...</previous\_action\_outcome>

<action\_description>...</action\_description>

<action>click("element\_id")</action>
```

Action space is provided below:

{{ action\_space }}

REASONING GUIDELINES (keep your chain‑of‑thought PRIVATE)

Always inspect the new screenshot (Current page) after every action.

Detect dropdowns/menus: If the clicked element now has expanded or you see
new list / menu items, assume a sub‑menu opened which could give you further options for the goa.

Avoid loops: if the visual state has not changed after two actions, rethink or quit.

Tasks can be ambiguous or slightly wrong; interpret generously. 
You can also take actions that could lead later to the goal.
Keep exploring until you think the goal has been reached.

</system\_prompt>

<user\_prompt>

\# Mini‑task

{{ goal\_object }}

Recent history (last steps, from oldest to newest. - n steps before.)

<action\_history>

  <step {{ h.idx }}>
  
    acted= <previous\_action{{ h.idx }}> {{ h.action }} </previous\_action{{ h.idx }}>

    saw  = <previous\_observation{{ h.idx }}> {{ h.observation }} </previous\_observation{{ h.idx }}>

  </step>

</action\_history>

Current page

URL: {{ open\_tabs }}

{{ state\_description }}

Analyze first the **Current page** content to understand very well what was the outcome of the last action and
what is the current state of the page.

**Reply with the following XML tags, according to the instructions and action space:**

```xml

<previous\_action\_outcome>...</previous\_action\_outcome>

<action\_description>...</action\_description>

<action>...</action>

```

</user\_prompt>

\end{tcolorbox}

\subsection{Agent exploration trajectory example}
\label{abs:trajectory_example}

\textbf{Partial Trajectory: Realizing Action Failure and Retrying}

\subsubsection*{Step 1: Agent Observes the Current UI State}

\begin{figure}[h]
    \centering
    \includegraphics[width=0.5\textwidth]{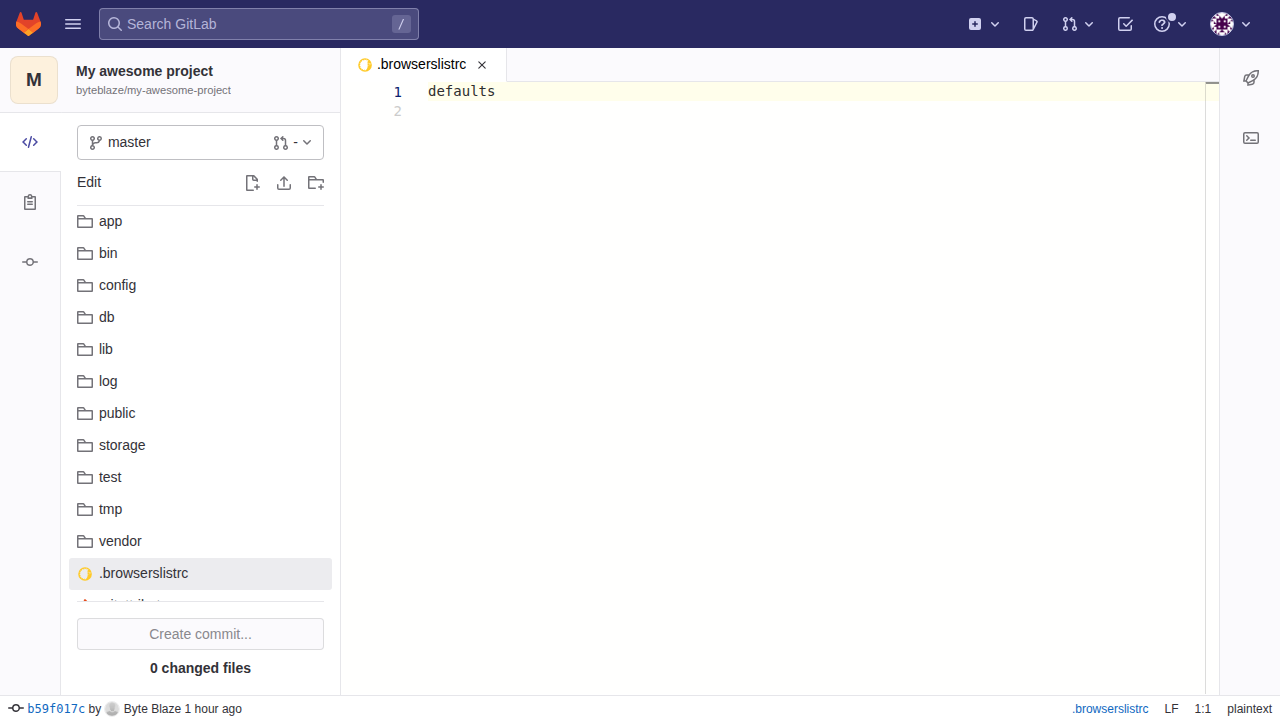}
    \caption{UI State: GitLab repository view with file tree and top navigation bar.}
\end{figure}

\begin{tcolorbox}[title=State Description and Macro-Actions, colback=gray!5, colframe=gray!80!black, breakable]
\footnotesize
\textbf{State:} The application is currently displaying a GitLab repository pa titled 'My awesome project'. It is showing the file structure on the left (e.g., app, bin, config, db, etc.), a file editor in the center displaying the contents of a file named '.browserslist', and a highlighted commit button ('Create commit...') at the bottom. The top navigation bar features search bar ('Search GitLab'), a project dropdown for branches or tags (currently 'master'), and various icons like settings, issues, and merge requests..

\textbf{Available Macro-Actions:}
\begin{itemize}
    \item "Interact with the branch/tag dropdown to explore branch-related features."
    \item "Click the 'settings' icon in the top navigation bar to access project settings."
    \item "Click the 'issues' icon to explore issue-related functionality."
    \item "Click the 'merge requests' icon to explore merge request-related functionality."
    \item "Click on a specific file or folder (e.g., 'app') in the file tree to navigate to its content."
    \item "Edit the text in the '.browserslistrc' file."
    \item "Interact with the commit history link in the bottom left corner to se previous commits."
    \item "Interact with the '...' menu next to the file name to explore additional file options."
    \item "Access the left sidebar toggle button to explore collapsing or expanding sidebar functionality."
    \item "Click the 'help' or question mark icon in the top navigation bar to explore help/documentation options."
\end{itemize}
\end{tcolorbox}

\subsubsection*{Step 2: Agent Selects a Macro-Action}

\begin{tcolorbox}[title=Chosen Macro-Action, colback=blue!5, colframe=blue!75!black, breakable]
\textbf{Goal:} Click the 'settings' icon in the top navigation bar to access project settings.
\end{tcolorbox}

\subsubsection*{Step 3: Agent Executes First Attempt}

\begin{tcolorbox}[title=Attempt 1, colback=white, colframe=black, breakable]
\textbf{Previous Outcome:} 
\begin{quote}
I am looking at a GitLab project page with a file browser interface. The top navigation bar contains several icons including notifications, merge requests, and settings.
\end{quote}

\textbf{Action Description:} 
\begin{quote}
I will click the settings/cog icon located in the top right corner of the navigation bar.\end{quote}

\textbf{Atomic Action:} \texttt{mouse\_click(1171, 23)}
\end{tcolorbox}

\begin{figure}[h]
    \centering
    \includegraphics[width=0.5\textwidth]{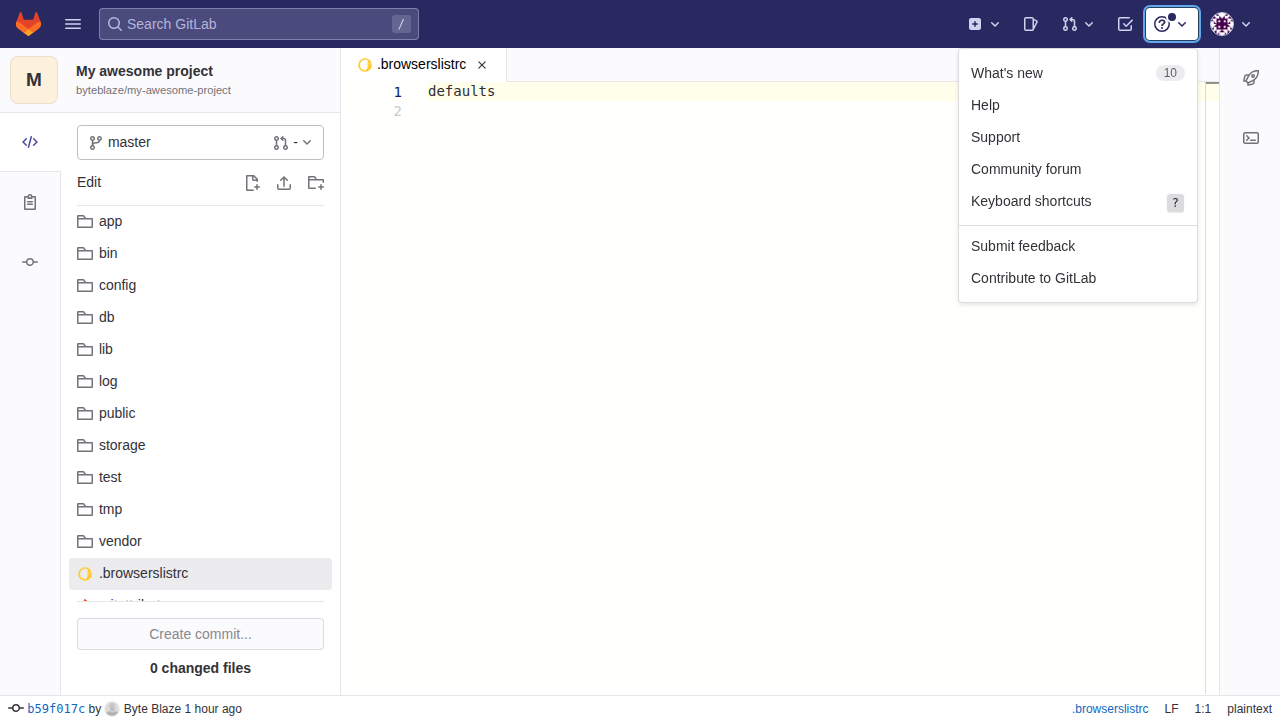}
    \caption{Post-action: The UI did not change, agent remains in repository view.}
\end{figure}

\subsubsection*{Step 4: Agent Recognizes Failure and Retries}

\begin{tcolorbox}[title=Attempt 2 – Retry, colback=white, colframe=red!70!black, breakable]
\textbf{Previous Outcome:} 
\begin{quote}
The previous click did not open the project settings. I'm still in the repository view showing the file browser interface.
There's a settings icon in the top navigation bar that may be in a different position than previously attempted.
\end{quote}

\textbf{Action Description:} 
\begin{quote}
I I will try clicking the settings icon (gear/cog icon) that appears to be located in the top navigation bar in a different position than previously attempted.
\end{quote}

\textbf{Atomic Action:} \texttt{mouse\_click(1170, 80)}
\end{tcolorbox}

\end{document}